\documentclass[11pt,reqno,final]{amsart}
\usepackage[font=small,margin=10pt,labelfont={bf},labelsep={space}]{caption}
\usepackage[scaled]{helvet}
\usepackage{fourier}
\usepackage{fancyhdr}
\usepackage{amsmath, amssymb, epsfig}
\usepackage{algorithmicx}
\usepackage[ruled]{algorithm}
\usepackage{algpseudocode}
\usepackage{wrapfig}
\usepackage[margin=1in]{geometry}
\usepackage{graphicx}
\usepackage{subfigure}
\usepackage{amsmath, amssymb, epsfig}
\usepackage{bm}
\usepackage{mathrsfs}
\usepackage{color}

\theoremstyle{plain}
\newtheorem{theorem}{Theorem}[section]

\theoremstyle{remark}
\newcommand{\vct}[1]{\bm{#1}}

\newcommand{\mtx}[1]{\bm{#1}}
\newcommand{\bigO}{\mathrm{O}}

\newcommand{\cut}{\operatorname{cut}}
\newcommand{\Ncut}{\operatorname{Ncut}}
\newcommand{\assoc}{\operatorname{assoc}}
\def\R{\mathbb{R}}

\graphicspath{{../Images/}}
\title{Spectral Clustering: An empirical study of Approximation Algorithms and its Application to the Attrition Problem
}
\thanks{Research performed at the Institute for Pure and Applied Mathematics (IPAM) in the University of California, Los Angeles (UCLA) during the Research in Industrial Projects for Students (RIPS) summer program of 2012.
B. Cung (bcung@ucla.edu) is with the University of California, Los Angeles. T. Jin (tonyjin1@stanford.edu) is with Stanford University. J. Ramirez (juan.ramirez.prado@itam.mx) is with the Instituto Tecnol\'ogico Aut\'onomo de M\'exico,
and A. Thompson (aubrey@huskers.unl.edu) is with the University of Nebraska-Lincoln.  C. Boutsidis (cboutsi@us.ibm.com) is with the IBM T.J. Watson Research Center and D. Needell (dneedell@cmc.edu) is with Claremont McKenna College.}

\author{ B. Cung, T. Jin, J. Ramirez\and A. Thompson \\\\ Advisors: C. Boutsidis \and D. Needell}

\small{\lhead{Cung, Jin, Ramirez, and Thompson}
\rhead{Approximation Algorithms for Spectral Clustering}
}
\begin{document}

\begin{abstract}
Clustering is the problem of separating a set of objects into groups (called clusters) so that objects within the same cluster are more similar to each other than to those in different clusters.  Spectral clustering is a now well-known method for clustering which utilizes the spectrum of the data similarity matrix to perform this separation.  Since the method relies on solving an eigenvector problem, it is computationally expensive for large datasets.  To overcome this constraint, approximation methods have been developed which aim to reduce running time while maintaining accurate classification.  In this article, we summarize and experimentally evaluate several approximation methods for spectral clustering. From an applications standpoint, we employ spectral clustering to solve  the so-called attrition problem, where one aims to identify from a set of employees those who are likely to voluntarily leave the company from those who are not. Our study sheds light on the empirical performance of existing approximate spectral clustering methods and shows the applicability of these methods in an important business optimization related problem.
%This paper summarizes several approximation techniques and analyzes their performance in terms of running time and accuracy in several application settings.  In particular, in the so-called attrition problem one aims to identify from a set of employees those who are likely to voluntirily leave the company from those who are not.  The spectral clustering method and its approximations are tested on this problem.  Experimental results indicate advantages and disadvantages of each approach, and identify optimal setting and sensitivity of various tuning parameters within each method.
\end{abstract}
\maketitle
%\begin{keywords}
%Spectral clustering, Approximation methods, Eigenvector approximation, Normalized cuts
%\end{keywords}
%
%\begin{AMS}
%62H30, 91C20, 91D30, 94C15
%\end{AMS}

%\pagestyle{myheadings}
%\thispagestyle{plain}
%\markboth{TEX PRODUCTION AND V. A. U. THORS}{SIAM MACRO EXAMPLES}

\section{Introduction}

%Motivate problem, discuss relation to Ncut problem

\begin{figure}[ht]
\centering
\subfigure[]{ \includegraphics[width=1.5 in,height=1.5 in] {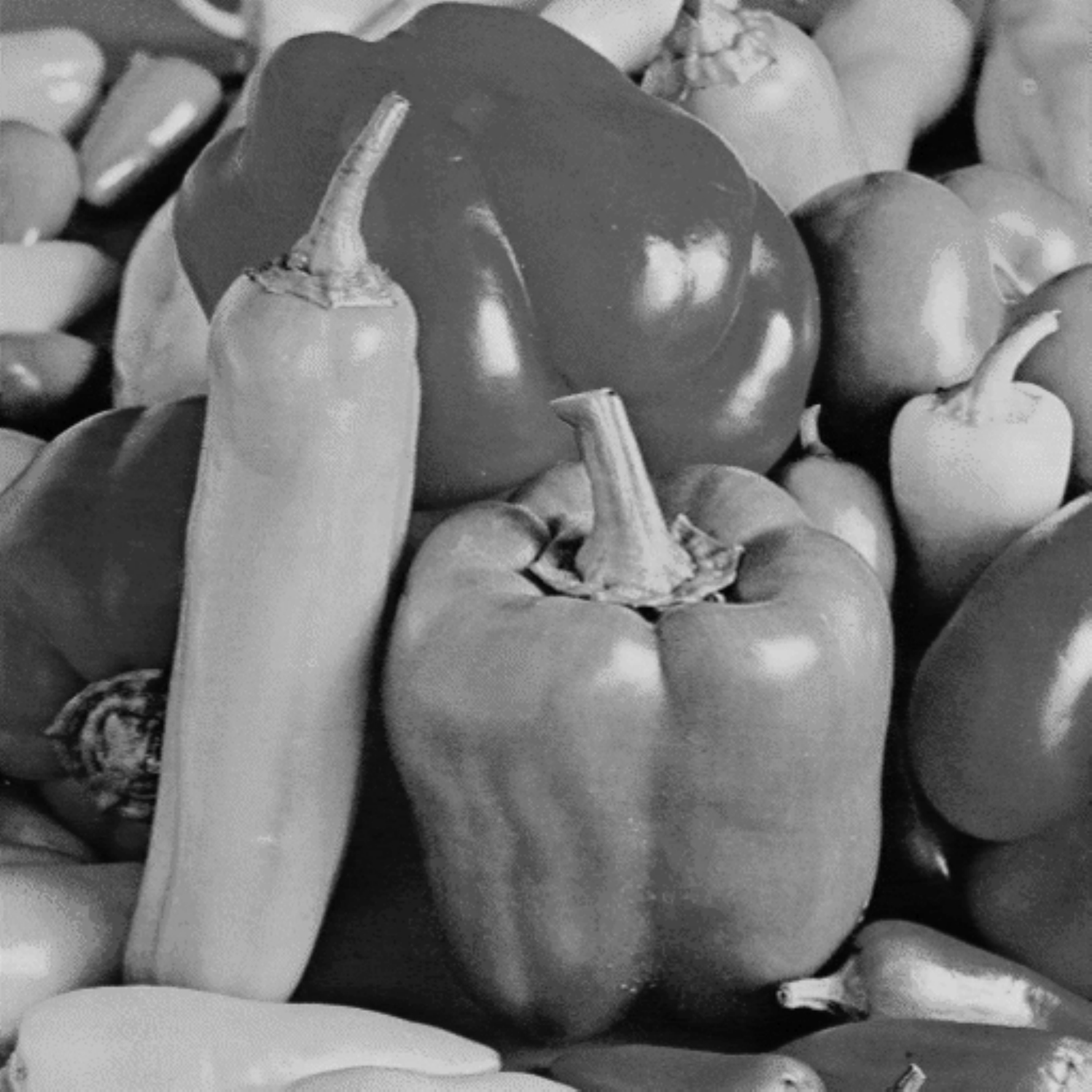}\hspace*{0.3in}
}
\subfigure[]{
   \includegraphics[width=1.5 in,height=1.5 in] {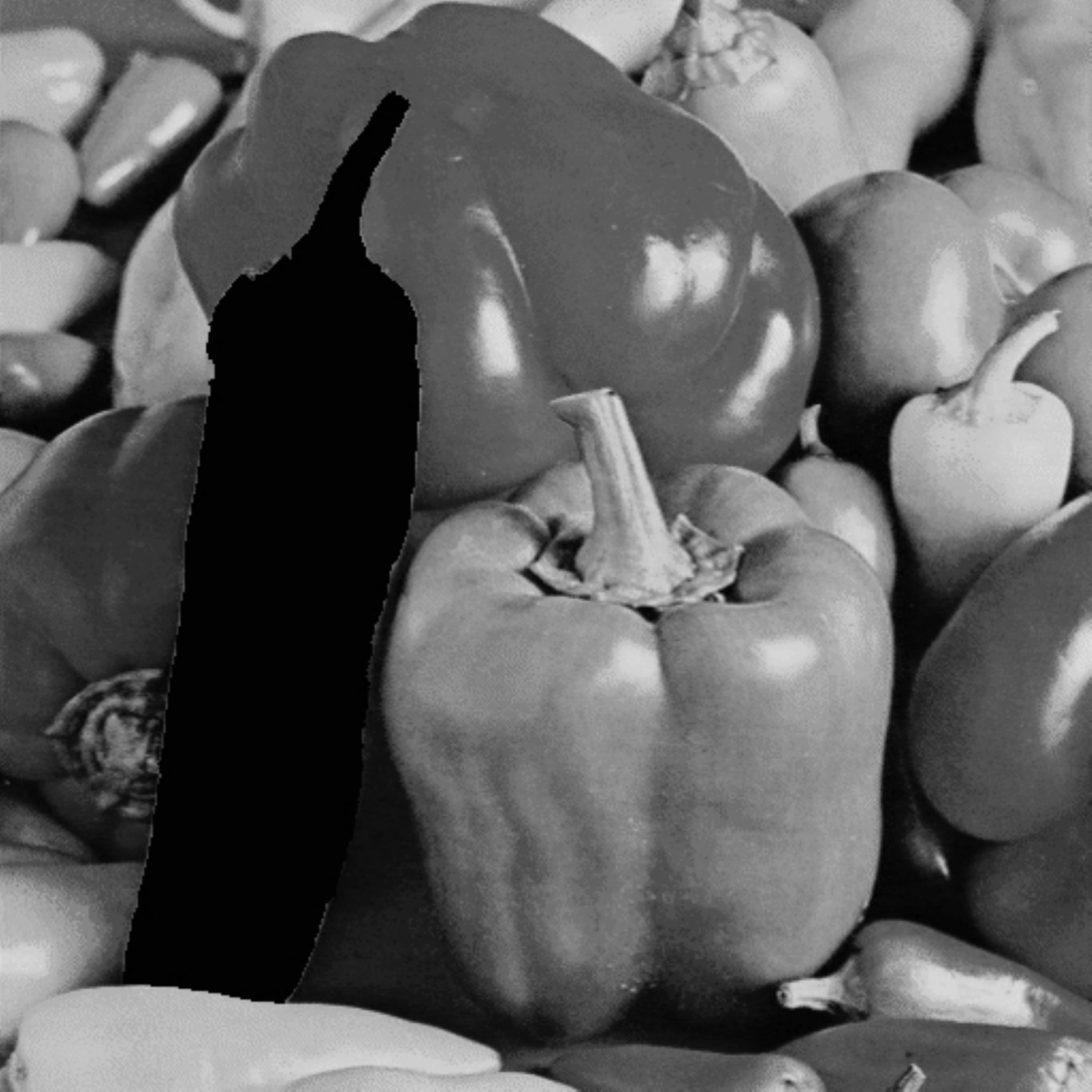}\hspace*{0.3in}
   }
   \subfigure[]{
      \includegraphics[width=1.5 in,height=1.5 in] {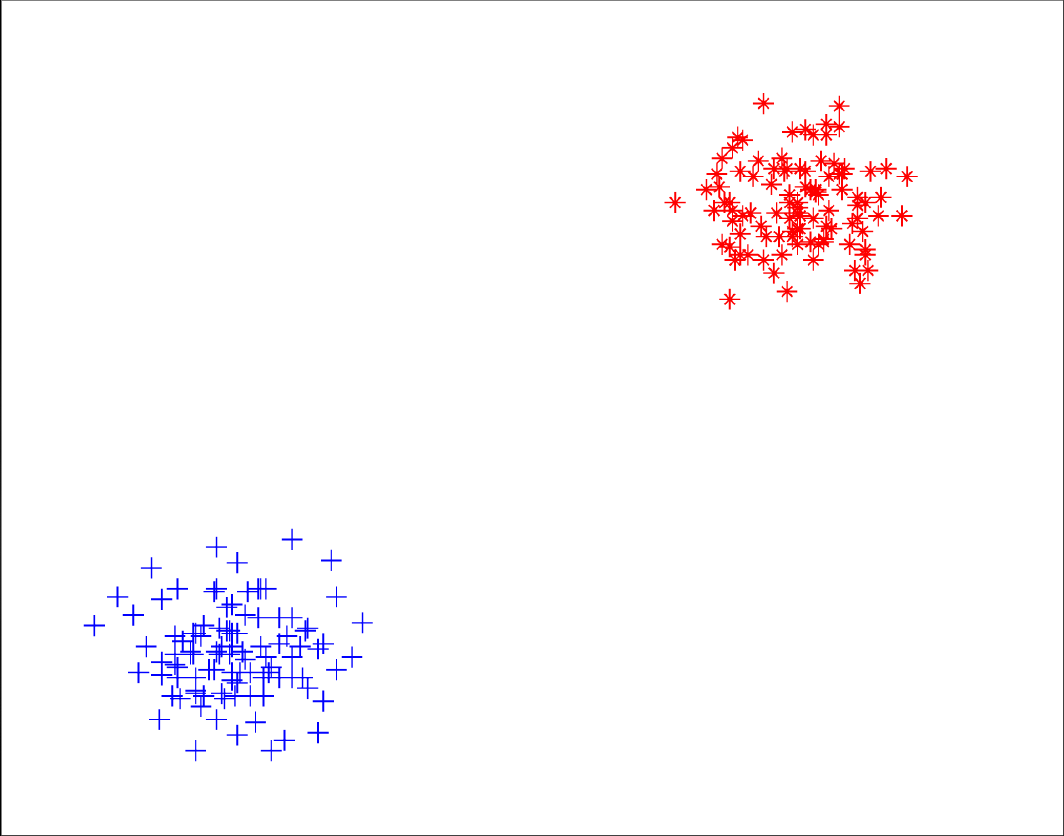}
     }
\caption{Examples of clustering: (a) original peppers image, (b) segmentation of peppers image, (c) two clusters in two dimensions. }\label{fig:examples}
\end{figure}

\textit{Clustering} or \textit{cluster analysis} addresses the problem of separating a set of objects into \textit{clusters} so that objects within each cluster are more similar to each other than to objects in different clusters.  The clustering problem has become ubiquitous in data mining and machine learning with applications ranging from image processing to bioinformatics.  What one means by clustering, and the type of clustering desired is application dependent.  For example, one may wish to segment an image such as that in Figure~\ref{fig:examples} (a)-(b).  In medical imaging, segmentation may aid in the identification of tumors, assist physicians in surgery and separate anatomical structures.  Computer vision applications utilize clustering methods to identify foreign objects in surveillance images or detect road signs for computer piloted vehicles.  In statistical analysis, the objects to be clustered may represent individuals in a population viewed as a vector of personal attributes.  For example, we will consider the \textit{attrition problem}: from a dataset of employees one wishes to identify which cluster of employees are likely to voluntarily leave the company and which are not.  With this problem as our overarching focus, we will consider here and throughout the case in which we wish to identify \textit{two} clusters in the data.  One can visualize this type of clustering in low dimensions, for example as seen in Figure~\ref{fig:examples} (c), where the ``correct'' cluster identification is obvious.

\subsection{Contributions}
As we discuss later in this section, there are different clustering algorithms such as $k$-means or spectral clustering.
The focus of this article is on \emph{spectral clustering}, a method which utilizes an eigenvector from the so-called data
similarity matrix. Computing eigenvectors of such matrices could be potentially a very expensive operation. Thus, faster
approximation algorithms for spectral clustering have appeared in the literature. The first contribution of this article
is to summarize and experimentally evaluate such approximation algorithms. Our second contribution is to apply spectral clustering
to a modern business optimization related problem which we call the \emph{attrition problem}: given a set of employees, we would like
to separate those who are likely to voluntarily resign from the company from those who are not. Such information could be of
tremendous value to the company because of the high costs to replace the workforce.  We present the empirical study of approximation
algorithms for spectral clustering in Section~\ref{sec:approx} and the case study to the attrition problem in Section~\ref{sec:attrition}.

\subsection{Clustering via $k$-means}
%Description, theoretical guarantees, drawbacks

The goal of clustering methods is to identify clusters automatically from the data input.  The \textit{$k$-means clustering} method is an approach that separates objects into $k$ clusters so that each object is assigned to the cluster whose \textit{mean} is nearest in the Euclidean sense~\cite{lloyd1982least,wu2009top}.  That is, given $n$ vectors $\vct{x_1}, \vct{x_2}, \ldots, \vct{x_n}$ in $d$-dimensional space, $\vct{x_j}\in\R^d$, the $k$-means method aims to minimize the sum of the squared intra-cluster distances:
$$
\sum_{i=1}^k\sum_{\vct{x_j}\in S_i}\|\vct{x_j} - \vct{\mu_i}\|_2^2,
$$
where, for $i=1,\dots,k,$ $S_i$ contains the indices of vectors in the $i$th cluster, and $\vct{\mu_i} \in \R^d$ denotes the mean (center) of vectors in that cluster.

Although this problem is in general NP-Hard~\cite{mahajan2009planar}, efficient iterative algorithms have been developed that often converge to a locally optimal solution (see e.g. Chapter 20 of~\cite{mackay2003information}).  Although variations in the method exist, the standard approach due to Lloyd (for $k=2$ clusters) consists of repeating the two steps described in Algorithm~\ref{alg:kmeans}.  We denote by $S^c$ the complement of the set $S$.

\begin{algorithm}{}
	\caption{$k$-means Clustering Method (for $k=2$)}\label{alg:kmeans}
\begin{algorithmic}[1]
\Procedure{}{$\vct{x_j}$'s, $\vct{\mu_1}$, $\vct{\mu_2}$, $T$}\Comment{data points $\vct{x_j}\in\R^d$, initial means $\vct{\mu_1}$, $\vct{\mu_2}$, number of iterations $T$}
%\State Compute a partition of $[n]$, call it $\mathcal{T}$. Assume it is a $(|\mathcal{T}|,\alpha,\beta)$ column paving of $\matA$.
\For {$t=1,2,\ldots, T$ }
\State Cluster by assigning each object to its closest mean:
\[
S_1 = \{\vct{x_j} : \|\vct{x_j} - \vct{\mu_1}\|_2 \leq \|\vct{x_j} - \vct{\mu_2}\|_2\}, \quad S_2 = S_1^c
\]
	\State Update the mean vectors:
	\[
	   \vct{\mu_1} = \frac{1}{|S_1|}\sum_{\vct{x_j}\in S_1}\vct{x_j}, \quad \vct{\mu_2} = \frac{1}{|S_2|}\sum_{\vct{x_j}\in S_2}\vct{x_j}
	\]
	\EndFor
\EndProcedure
\end{algorithmic}
\end{algorithm}

\begin{figure}[ht]
\centering
  \subfigure[]{ \includegraphics[width=1.5 in,height=1.5 in] {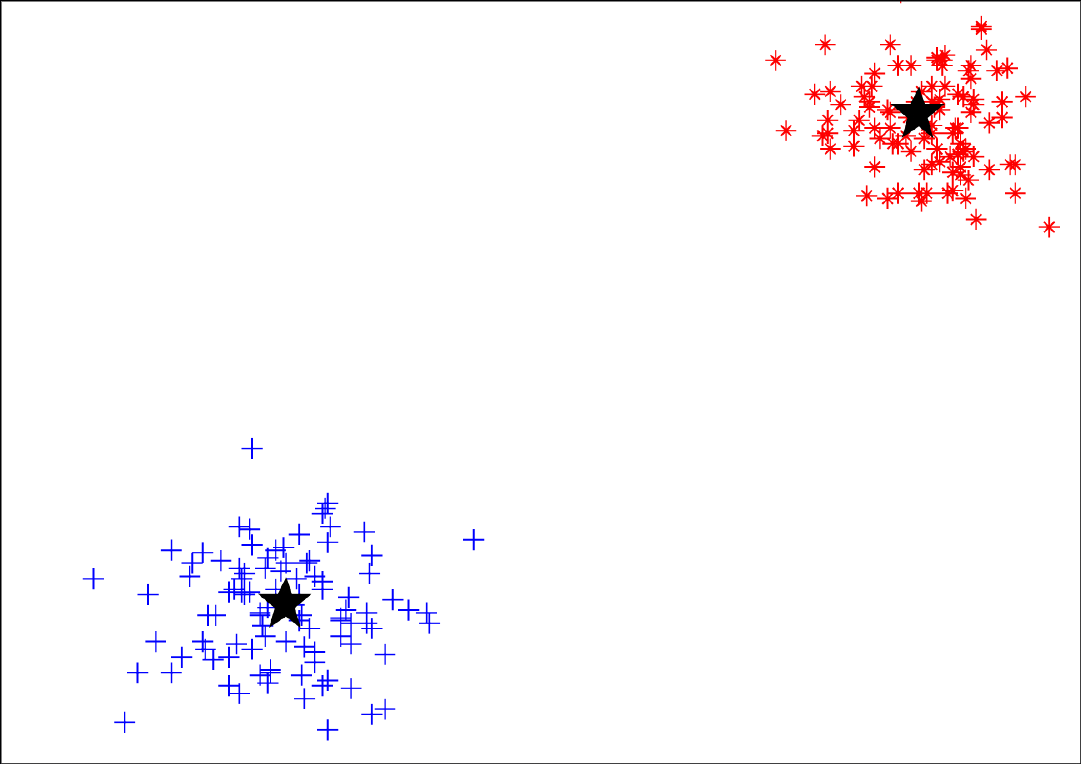}\hspace*{0.3in}
  }
  \subfigure[]{
      \includegraphics[width=1.5 in,height=1.5 in] {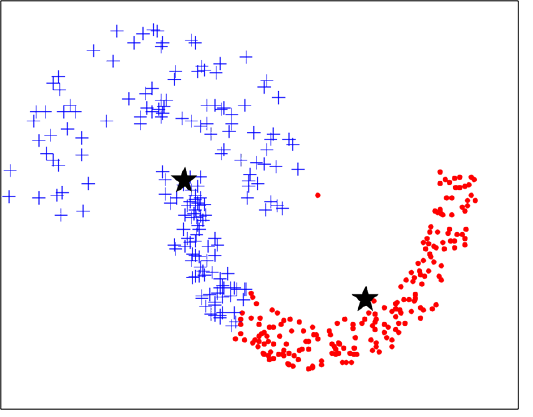}
      }
\caption{The $k$-means clustering method: (a) two clusters in two dimensions with cluster means converged to cluster centers (marked with stars); (b) non-spherical clusters are difficult to identify via $k$-means clustering. }\label{fig:kmeans}
\end{figure}

To separate the points into more than $2$ clusters, one extends the method for $k$-means in the natural way.  The runtime of the method is $\bigO(knT)$.  When the mean of each cluster converges toward the true cluster center, the $k$-means method performs well.  This is the case, for example, when the clusters are each of similar size and have a spherical shape as seen in Figure~\ref{fig:kmeans} (a).  However, when the clusters are not linearly separable, as in Figure~\ref{fig:kmeans} (b), $k$-means may often incorrectly assign points to clusters.  Although $k$-means performs well in many settings, there are also applications where these limitations are apparent, and this leads us to search for other methods that will work for more general purposes.

\subsection{Spectral Clustering}

An alternative way to approach the clustering problem is to view the data points as a graph.  Each vertex of the graph will represent a data point, and each edge will represent the \textit{similarity} between the two corresponding vertices.  To that end, for $n$ data points $\vct{x_1}, \vct{x_2}, \ldots, \vct{x_n}$ in $d$-dimensional space, denote by $\mtx{X}$ the $n\times d$ data matrix whose rows contain the data vectors $\vct{x_j}^T$.  We construct a \textit{similarity matrix} $\mtx{W} \in \R^{n \times n}$ whose $(i,j)$th entry gives the similarity between the two corresponding data points:

\begin{equation}\label{eq:W}
\mtx{W}_{ij}=\exp\left(-{\frac{\|\vct{x_i}-\vct{x_j}\|^2}{\sigma_{ij}^2}}\right),
\end{equation}

where $\sigma_{ij}$ is a tuning parameter to be chosen later.  The similarity matrix $\mtx{W}$ induces a complete graph $(V,E,\mtx{W})$ where $V$ is the set of vertices (objects) to be clustered, $E$ is the set of edges, and $\mtx{W}$ represents the weights of the edges.  The clustering problem can then be viewed as the partitioning of the graph into sets of vertices such that the edges within the sets have large weights, and the edges across sets have small weights.  Formally, in the $2$-clustering setting, we wish to identify sets $A$ and $B$ which minimize the so-called \textit{normalized cut} objective,

\[\Ncut(A, B) = \frac{\cut(A, B)}{\assoc(A, V)} + \frac{\cut(A, B)}{\assoc(B, V)},\]
where the cut and association functions are defined by
\[\cut(A, B) = \sum_{\substack{\vct{x_i} \in A \\ \vct{x_j} \in B}} \mtx{W}_{ij}, \quad\text{}\quad \assoc(A, V) = \sum_{\substack{\vct{x_i} \in A \\ \vct{x_j} \in V}} \mtx{W}_{ij},
\quad\text{and} \quad \assoc(B, V) = \sum_{\substack{\vct{x_i} \in B \\ \vct{x_j} \in V}} \mtx{W}_{ij}.
\]

\begin{figure}[ht]
\centering
  \subfigure[]{ \includegraphics[width=2 in] {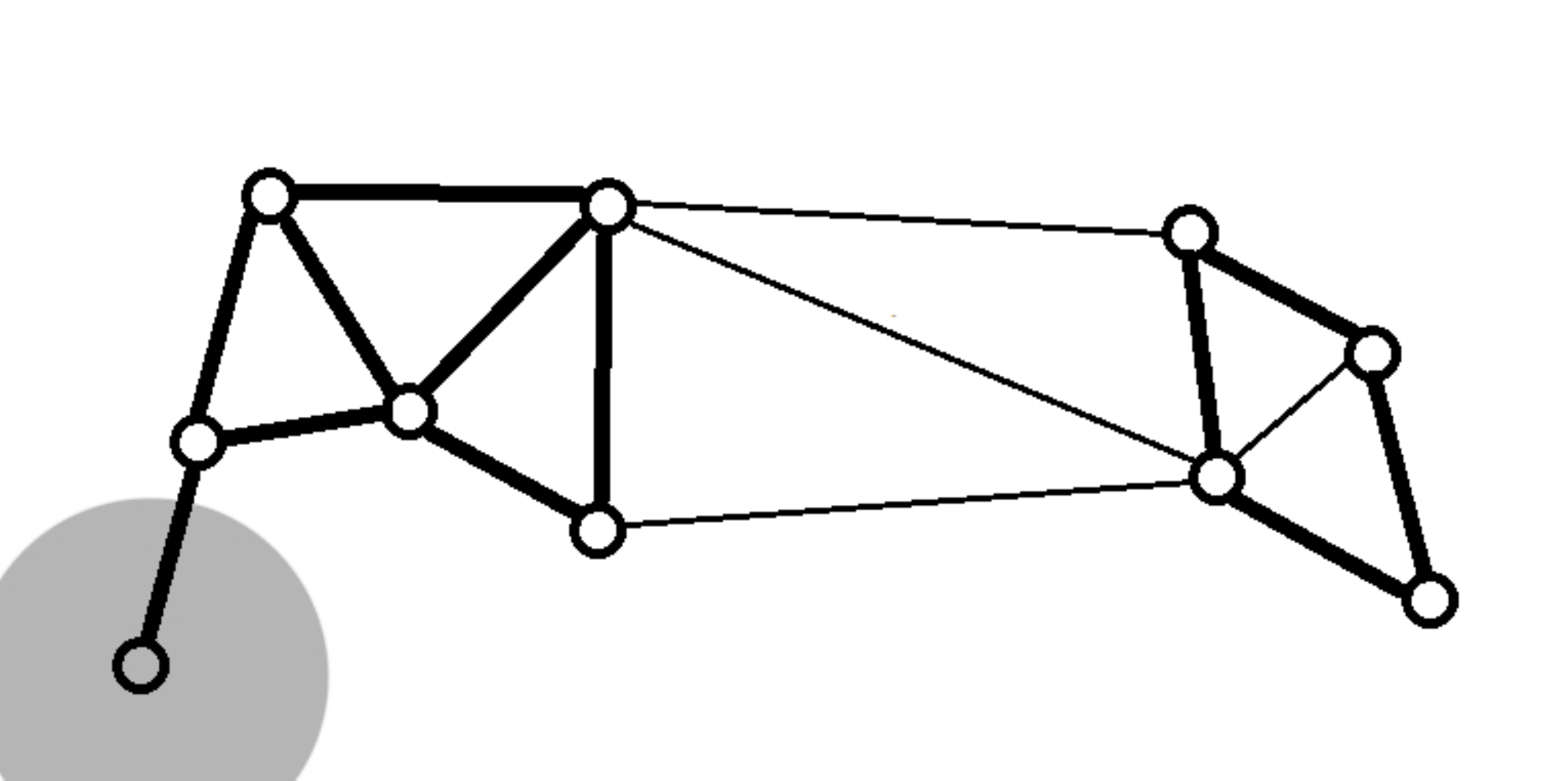}\hspace*{0.3in}}
     \subfigure[]{ \includegraphics[width=2 in] {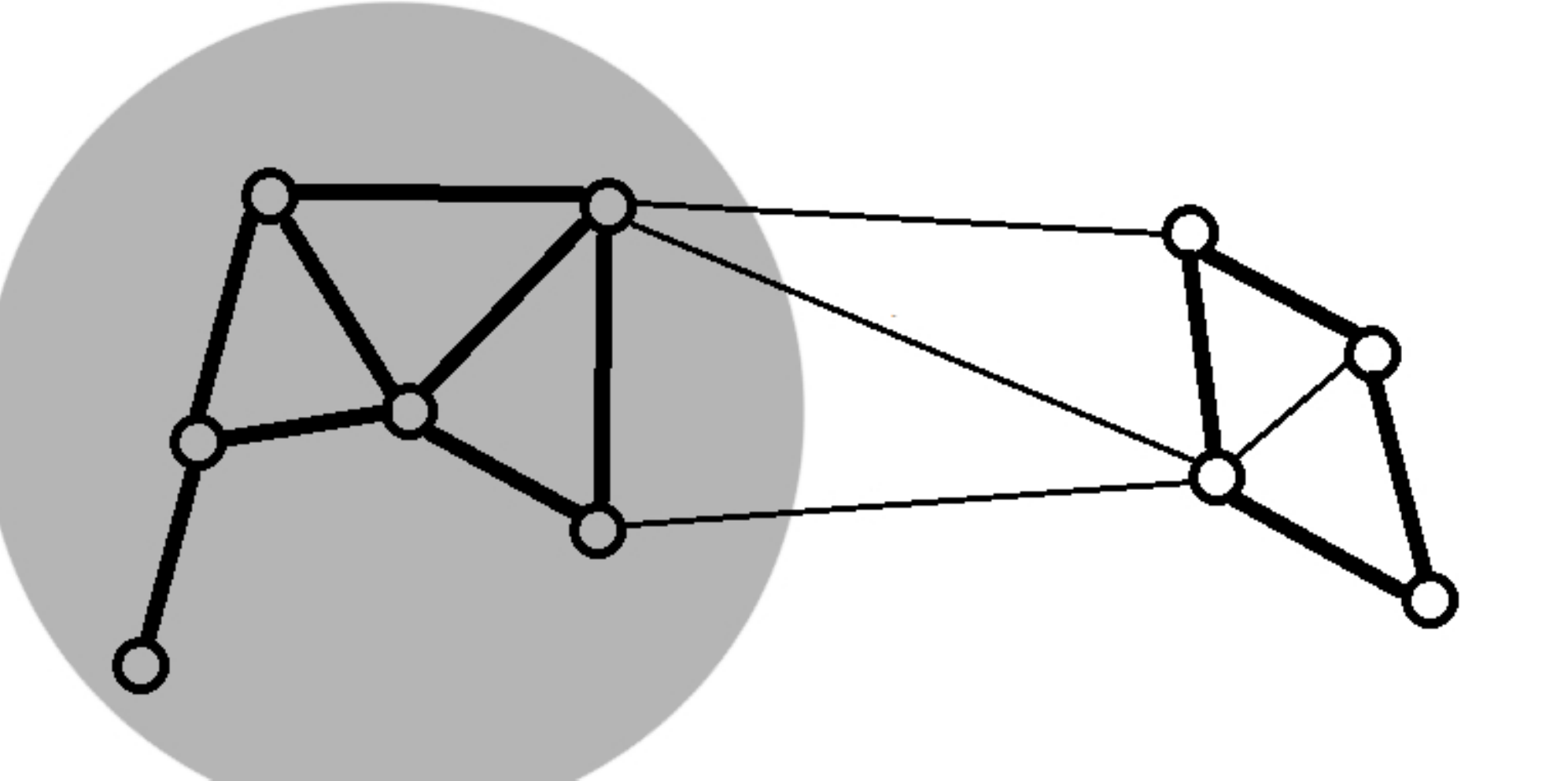}}
\caption{Two examples of graph partitioning (one set is shown shaded and the other unshaded): (a) Minimizing the cut of the graph, (b) minimizing the normalized cut of the graph. }\label{fig:graphs}
\end{figure}

The numerators of $\Ncut$ defined in this way guarantee that the weights between the clusters $A$ and $B$ are small.  On the other hand, if we simply minimized the cut function, one might obtain cuts for which $A$ is a very small set of vertices (perhaps even just one vertex) and $B$ is the remaining vertices, as shown in Figure~\ref{fig:graphs} (a).  To avoid these trivial cuts, we divide by the association function, which sums the weights between a set of vertices and \textit{all} nodes.  If a set of vertices in the partition is too small, its association will be small, leading to a large $\Ncut$.  With this normalization, one hopes to avoid this type of bias and obtain cuts as in Figure~\ref{fig:graphs} (b).

\begin{figure}[ht]
\centering
  \includegraphics[width=2.0 in] {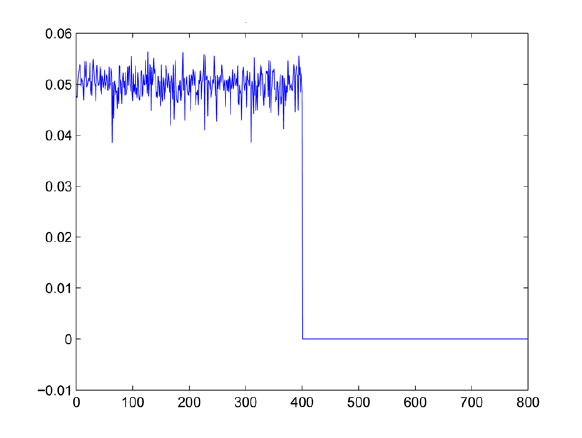}
\caption{An example of an eigenvector obtained from the spectral clustering method (horizontal axis represents the index, vertical the value of the eigenvector at that index). Here we assign the first $400$ objects to one cluster and the second $400$ to the other.}\label{fig:eigenvector}
\end{figure}

Minimizing the normalized cut is NP-Complete in general (see~\cite{shi2000normalized} for the proof; originally due to Papadimitriou).  However, recently a relaxation of the optimization problem has been reduced to an eigenvector problem~\cite{shi2000normalized}.  Given the $n\times n$ similarity matrix $\mtx{W}$, one defines the normalized Laplacian\footnote{Note that one may also consider the Laplacian $\mtx{L} = \mtx{D} - \mtx{W}$.} matrix $\mtx{L} \in \R^{n \times n}$ by
\begin{equation}\label{eq:L}
\mtx{L} = \mtx{D}^{-1/2} (\mtx{D} - \mtx{W})\mtx{D}^{-1/2},
\end{equation}
where $\mtx{D} \in \R^{n \times n}$ is the diagonal matrix of degree nodes,
\begin{equation}\label{eq:D}
\mtx{D}_{ii}=\sum_j \mtx{W}_{ij}.
\end{equation}
 Shi and Malik argued that the eigenvector corresponding to the second smallest eigenvalue of $\mtx{L}$ corresponds to a linear transformation of the relaxed solution to the $\Ncut$ problem~\cite{shi2000normalized}.  Indeed, the clustering is then performed by selecting an appropriate threshold, and assigning indices of the eigenvector with large values to one cluster, and indices with small values to the other.  For example, with the eigenvector plotted in Figure~\ref{fig:eigenvector}, the first $400$ data points would be assigned to one cluster and the second $400$ to the other.  This gives rise to the following formal definition of the spectral clustering algorithm.

\begin{algorithm}{}
	\caption{Spectral Clustering Method (for two clusters)}\label{alg:SC}
\begin{algorithmic}[1]
\Procedure{}{$\mtx{X}$, $\sigma$}\Comment{$n\times d$ data matrix $\mtx{X}$, tuning parameter $\sigma$}
%\State Compute a partition of $[n]$, call it $\mathcal{T}$. Assume it is a $(|\mathcal{T}|,\alpha,\beta)$ column paving of $\matA$.
\State Construct the similarity matrix $\mtx{W}$ in~\eqref{eq:W}, degree matrix $\mtx{D}$ in~\eqref{eq:D} and Laplacian $\mtx{L}$ in~\eqref{eq:L}

	\State Compute the eigenvector corresponding to the second smallest eigenvalue of $\mtx{L}$
	
	\State Assign the indices in the eigenvector with large values to one cluster, the rest to the other
\EndProcedure
\end{algorithmic}
\end{algorithm}

The step which is most computationally burdensome is the eigenvector computation.  In general this step yields an $\bigO(n^3)$ running time.  This cost is often detrimental for large applications and is one of the biggest drawbacks to spectral clustering methods.

\begin{figure}[ht]
\centering
\subfigure[]{
   \includegraphics[width=1.5in,height=1.5in] {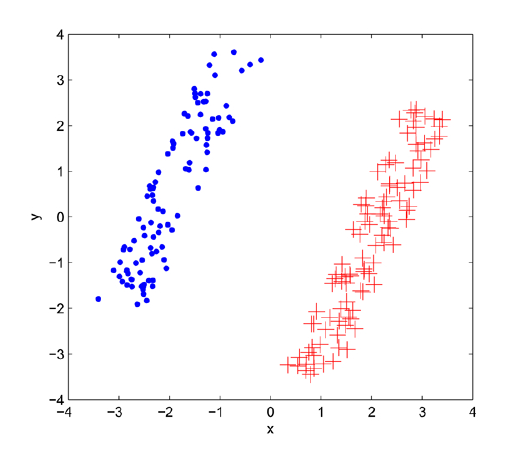}
 }
\subfigure[]{
   \includegraphics[width=1.5in,height=1.5in] {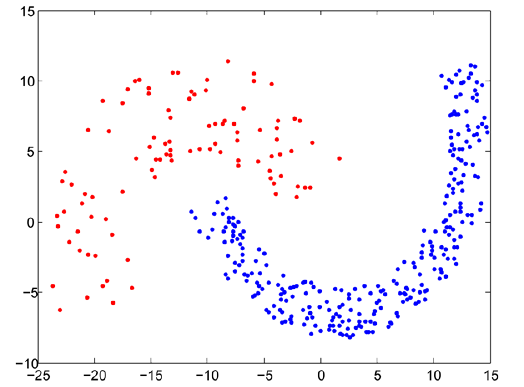}
 }
\subfigure[]{
   \includegraphics[width=1.5in,height=1.5in] {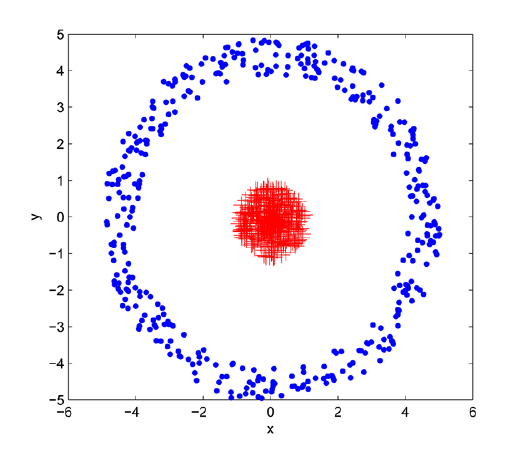}
 }
\\
\hspace{15mm}\subfigure[]{
   \includegraphics[width=1.5in,height=1.5in] {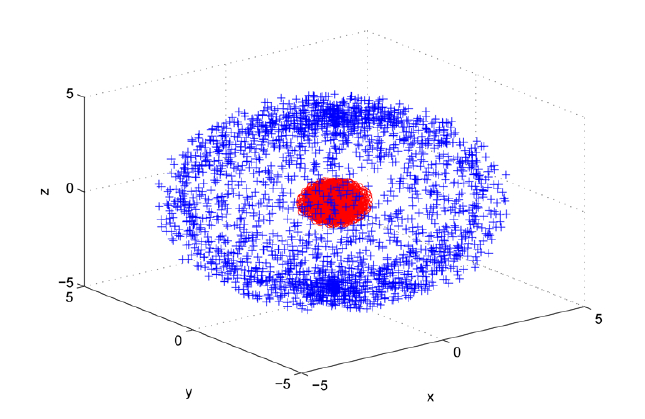}
 }
\subfigure[]{
   \includegraphics[width=1.5in,height=1.5in] {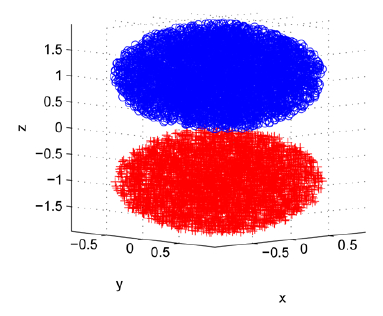}
 }
 \subfigure[]{
   \includegraphics[width=1.5in,height=1.5in] {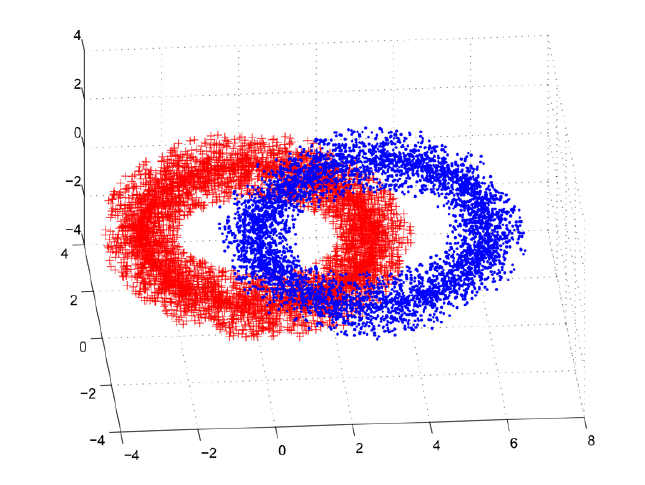}
 }
\caption{Accurate clustering results for data sets of many shapes and sizes using spectral clustering: (a) Gaussian strips, (b) Interlocked half rings, (c) Concentric rings, (d) Concentric spheres, (e) Tangent spheres, and (f) Interlocked rings.}
\label{fig:SCresults}
\end{figure}

Experimental results using the spectral clustering method are shown in Figure~\ref{fig:SCresults}.  Here we use the self-tuning approach by Zelnik-Manor and Perona \cite{perona2004self} for obtaining the similarity matrix.  Consider the vector in $\mathbb{R}^{n}$ defined entrywise by
\begin{equation}\label{eq:nu}
\nu_i=\|\vct{x_i}-\vct{x_{i_K}}\|_2,
\end{equation}
where $\vct{x_{i_K}}$ denotes the $K^{th}$ closest neighbor to $\vct{x_i}$.  We then set the scaling parameter $\sigma_{ij}$ as
\begin{equation}\label{eq:selftune}
\sigma_{ij} = \nu_i\nu_j.
\end{equation}
As in the article, we set $K=7$ for all our experiments with the spectral clustering algorithm.  The experiment performed on the interlocked rings, interlocked half rings, and Gaussian strips were run on Intel Core 2 Duo E8500 3.16 GHz machines with 6 MB cache and 16 GB memory.  The concentric spheres and concentric rings experiments were run on Intel Core i7 870 2.93 GHz machines with 4 cores, 8 MB cache and 16 GB memory.  The tangent spheres experiments were run on an Intel Xeon W3520 2.67 GHz machine with 4 cores, 8 MB cache and 16 GB memory.

 The running times and dataset sizes are summarized in Table~\ref{tbl:SCresults}.

\begin{table}[ht]
\begin{tabular}{|l||l|l|}
\hline
Dataset & Size $n$ & Running time (s)\\
\hline
Gaussian strips & 200 & 0.17\\
\hline
Interlocked half rings & 373 & 0.73\\
\hline
Concentric rings & 800 & 8.02\\
\hline
Concentric spheres & 5000 & 855.12\\
\hline
Tangent spheres & 10000 & 6365.4\\
\hline
Interlocked rings & 10000 & 6870.7\\
\hline
\end{tabular}

\caption{Running times and dataset sizes for problems of Figure~\ref{fig:SCresults}.}
\label{tbl:SCresults}
 \end{table}

 Although the spectral clustering method performs accurate cluster identification even for datasets challenging to the $k$-means method, its cubic runtime is a practical obstacle in many applications.  For this reason, approximation methods have been developed which efficiently approximate the spectrum of the Laplacian matrix $\mtx{L}$.  In this article, we consider four popular approximation methods.  The \textit{Fast spectral clustering} method~\cite{yan2009fast} and Extensible spectral clustering method~\cite{wang2009approximate} both identify a small set of representative points from the dataset, perform spectral clustering on this much smaller set of points, and extend the identification to the remaining data points.  An alternative way to reduce the dimension of the similarity matrix is to randomly sample values from the matrix to obtain a smaller submatrix, which is the basis of the \textit{Spectral clustering on a budget}~\cite{shamirspectral} and Nystr\"{o}m  methods~\cite{fowlkes2004spectral}.

 \subsection{Organization}  The remainder of the article is organized as follows.  The four approximation methods are described and discussed in Section~\ref{sec:approx}.  Section~\ref{sec:experiments} displays numerical results used for a comparison between the methods.  In Section~\ref{sec:attrition} we focus on the attrition problem and analyze how each method performs at that task.  We conclude in Section~\ref{sec:discuss} with a discussion of the findings.

\section{Approximation Methods}\label{sec:approx}

The fundamental idea behind efficiently approximating the spectral clustering method is to reduce the problem size to be clustered.  To maintain accurate cluster identification, one hopes that the reduced problem preserves the same cluster structure as the original problem.  The two main approaches to this goal that we discuss here rely either on randomness to reduce the dimension, or some preprocessing algorithmic step to ensure that the smaller set is a good representation of the original. To evaluate accuracy of the method one uses the results of spectral clustering as the ground truth, and compares the output of the other methods to that.  To evaluate in a general sense, rather than example to example, one may wish to compare the eigenvector computed by the approximation method to that of the spectral clustering method.  We discuss these notions and describe the methods in the remainder of this section.

%\subsection{Spectral Clustering on Perturbed Data}
The common theme between approximate spectral clustering methods is that the $n\times n$ similarity matrix $\mtx{W}$ is downsampled so that clustering can be performed efficiently.  Such downsampling will of course lead to errors in the computed eigenvectors, and one wishes to quantify the magnitude of such perturbations to validate the accuracy of the approximate method.  In a general context, we can view this process as the perturbation of the Laplacian) matrix $\widetilde{\mtx{L}} = \mtx{L} + \mtx{E}$ where $\mtx{E}$ is some error matrix and $\widetilde{\mtx{L}}$ is the perturbed Laplacian matrix.
Standard results from linear algebra guarantee the following bound on the perturbation of eigenvectors.

\begin{theorem}[Eigenvector Perturbations~\cite{yan2009fast,huang2008spectral,stewart1973introduction}]
Suppose $\widetilde{\mtx{L}} = \mtx{L} + \mtx{E}$ and denote by $\widetilde{\vct{v}_i}$ and $\vct{v}_i$ the $i$th eigenvectors of $\widetilde{\mtx{L}}$ and $\mtx{L}$, respectively, corresponding to the $i$th smallest eigenvalue. Then
$$
\|\widetilde{\vct{v}_2} - \vct{v}_2\|_2 \leq \frac{1}{\lambda_2 - \lambda_3}\|\mtx{E}\| + \bigO(\|\mtx{E}\|^2),
$$
where $\lambda_i$ denotes the $i$th smallest eigenvalue of $\mtx{W}$.
\end{theorem}

This result shows that the perturbation in the eigenvectors is controlled by the (spectral) norm of the perturbation in the matrix, and the \textit{eigengap} $\lambda_2 - \lambda_3$.  This theory can be extended to bound the angles between eigenspaces of the original and perturbed matrices as well as the norm of their projections~\cite{hunter2010performance}.

In analyzing approximation methods, one wishes to determine the tradeoff between accuracy and efficiency.
To quantify theoretically and empirically the performance of an approximation method, we define the mis-clustering rate by

\begin{equation}\label{eq:rho}
\rho = \frac{1}{n}\sum_{i=1}^n \mathbb{1}_i,
\end{equation}

where $\mathbb{1}_i$ is the indicator function which is equal to $1$ if object $\vct{x}_i$ is clustered correctly and zero otherwise.  We assume here that the correct clustering is given by the spectral clustering method, and compare the results of the approximation methods against that standard.  One can then bound the mis-clustering rate by the difference in the eigenvectors under certain assumptions.

\begin{theorem}[Misclustering rate~\cite{huang2008spectral,yan2009fast}]\label{thm:rho}
Suppose $\widetilde{\mtx{L}} = \mtx{L} + \mtx{E}$ and denote by $\widetilde{\vct{v}_2}$ and $\vct{v}_2$ the $2$nd (smallest) eigenvectors of $\widetilde{\mtx{L}}$ and $\mtx{L}$, respectively.  Then when both sets of eigenvectors partition the data into two clusters and the perturbations in the eigenvectors satisfy the componentwise assumptions of~\cite{huang2008spectral},
$$
\rho \leq \|\widetilde{\vct{v}_2} - \vct{v}_2\|_2^2.
$$
\end{theorem}

This result motivates the development of approximation methods which yield small perturbations in the eigenvectors of the downsampled matrix.

\subsection{Fast Spectral Clustering}
The \emph{fast spectral clustering algorithm} by Yan et al.~\cite{yan2009fast} consists of two major parts:  data preprocessing and spectral clustering.  The goal of the data preprocessing is to construct a smaller, but representative set of points to undergo spectral clustering rather than the original large dataset.  Since the $k$-means method itself identifies $k$ representative points (usually the cluster means), it seems a natural way to identify a representative set of points even if $k$ is larger than the number of clusters.  One can then perform spectral clustering efficiently on the representative points, and assign clusters to the entire original dataset by simply choosing the cluster containing the closest representative point.  Indeed, the algorithm for fast spectral clustering is described in Algorithm~\ref{alg:FSC} below.

\begin{algorithm}{}
	\caption{Fast Spectral Clustering Method}\label{alg:FSC}
\begin{algorithmic}[1]
\Procedure{}{$\mtx{X}$, $k$, $T$}\Comment{$n\times d$ data matrix $\mtx{X}$, number of representative points $k$, number of iterations $T$}
%\State Compute a partition of $[n]$, call it $\mathcal{T}$. Assume it is a $(|\mathcal{T}|,\alpha,\beta)$ column paving of $\matA$.
\State Find $k$ representative points (centroids $\vct{y_1},...,\vct{y_k}$) via $k$-means

	\State Create a correspondence table that associates each $\vct{x_i}$ with the nearest cluster centroid $\vct{y_j}$;
	
	\State Run spectral clustering on the data matrix $\mtx{Y}$ of centroids to obtain a clustering assignment
	
	\State Use the correspondence table to recover cluster membership for each point $\vct{x_i}$
\EndProcedure
\end{algorithmic}
\end{algorithm}

The complexity for the $k$-means step is $\bigO(knT)$, and since spectral clustering is only run on the $k$ representative points, that step yields a cost of just $\bigO(k^3)$.  The remaining assignment steps cost at most $\bigO(n)$, yielding an overall runtime of $\bigO(knT + k^3)$.  This is of course a significant improvement over the cubic $\bigO(n^3)$ of spectral clustering when $k$ and $T$ are chosen small enough.  To quantify precisely this tradeoff between efficiency and accuracy, the perturbation theory of Theorem~\ref{thm:rho} can be utilized.  Indeed, results on fast spectral clustering guarantee the following bound on the mis-clustering rate.

\begin{theorem}[Spectral misclustering rate~\cite{yan2009fast}]
Assume that the assumptions of Theorem~\ref{thm:rho} hold.  Then the mis-clusterting rate $\rho$~\eqref{eq:rho} for fast spectral clustering satisfies
$$
\rho \lesssim \frac{2}{(\lambda_2 - \lambda_3)^2} \|\mtx{L} - \widetilde{\mtx{L}}\|_F^2,
$$
where $\|\cdot\|_F^2$ denotes the Frobenius norm, $\mtx{L}$ and $\widetilde{\mtx{L}}$ denote the Laplacian and perturbed Laplacian, and the symbol $\lesssim$ implies that higher order terms are ignored in the relation.
\end{theorem}

This result demonstrates that the mis-clustering rate is again controlled by the eigengap and the perturbations in the Laplacian incurred via fast spectral clustering.  The latter term can be bounded in special cases, see~\cite{yan2009fast} for details.

%\notate{Talk about theoretical guarantees. Advantages, disadvantages. Specific experimental results?}

\subsection{Extensible Spectral Clustering}

The notion of identifying a small representative sample of the data on which to initially perform spectral clustering can also be generalized.  This class of methods are given the name \textit{extensible spectral clustering} (eSPEC)~\cite{pavan2005efficient,bezdek2006approximate,wang2009approximate}.  Here, one performs spectral clustering on the representative sample of the data, and assign each object in the original dataset to cluster based on its $m$ closest neighbors within the representative sample.  We again use the similarity matrix~\eqref{eq:W} to measure "closeness."  The general model is described in Algorithm~\ref{alg:espec}.

\begin{algorithm}{}
	\caption{Extensible Spectral Clustering Method}\label{alg:espec}
\begin{algorithmic}[1]
\Procedure{}{$\mtx{X}$, $m$, $S$}\Comment{$n\times d$ data matrix $\mtx{X}$, neighboring parameter $m$, representative sample $S$}
%\State Compute a partition of $[n]$, call it $\mathcal{T}$. Assume it is a $(|\mathcal{T}|,\alpha,\beta)$ column paving of $\matA$.
\State Run spectral clustering on the representative sample $S$ to obtain a clustering assignment

	%\State Create a correspondence table that associates each $\vct{x_i}$ with the nearest cluster centroid $\vct{y_j}$;
	
	\State For each object $i$ in $S^c$, find its $m$ closest neighbors in $S$
	
	\State Assign each object $i$ to the cluster containing the majority of its $m$ closest neighbors
\EndProcedure
\end{algorithmic}
\end{algorithm}

There are of course many ways one can initially obtain the representative sample.  As in the fast spectral clustering method, one can utilize the $k$-means method to identify a good representative sample $S$.  Indeed, if the centroids found by the $k$-means method coincide with data points in the set, the extensible spectral clustering method with $m=1$ is the same as the fast spectral clustering method.  Alternatively, the representative sample can be chosen randomly.  For example, one can simply sample uniformly at random from the dataset (see e.g.~\cite{pavan2005efficient,talwalkar2008large,wang2009approximate} and references therein) or according to some other probability distribution such as one that assigns probabilities proportional to the norms of each column~\cite{drineas2006fast}.  In the experiments section below, we see that using uniform sampling with just $m=1$ provides accurate results even for reasonably small sample sizes.  In this case the running time of the method is dominated by the size of the sample, $\bigO(|S|^3)$.

%\notate{Theory? Runtime? Advantages/disadvantages? Other approaches? Specific experimental results?}

\subsection{Nystr\"om Method}
Both the fast spectral clusterting method and extensible spectral clustering reduce the dimension of the clustering problem by subsampling the \textit{objects} in the data.  An alternative approach is to subsample the similarity matrix $\mtx{W}$~\eqref{eq:W} itself.  In this case, one uses a submatrix of $\mtx{W}$ and asks that the submatrix approximates the entire matrix $\mtx{W}$ well.  This is the motivation behind the \textit{Nystr\"om method}~\cite{nystrom1930praktische,baker1977numerical,fowlkes2004spectral}.

To that end, we decompose the $n \times n$ similarity matrix $\mtx{W}$ so that
\begin{equation}\label{eq:decomposeW}
\mtx{W} = \left( \begin{array}{ccc} \mtx{W_{1 1}} & \mtx{W_{2 1}}^T \\
             \mtx{W_{2 1}} &  \mtx{W_{2 2}} \end{array} \right),
\end{equation}
where $\mtx{W_{11}} \in \mathbb{R}^{m\times m}, \mtx{W_{21}} \in \mathbb{R}^{(n-m)\times m}$, and $\mtx{W_{22}} \in \mathbb{R}^{(n-m)\times (n-m)}$.  Choosing $m\ll n$, $\mtx{W_{22}}$ is very large, and this is thus the part we wish to approximate.

To do so, one computes the similarity matrix for only the $m$ sampled data points, represented by $\mtx{W_{1 1}}$.  The relationship between the sampled data points and the rest of the points is given by $\mtx{W_{2 1}}$.  Then only $\mtx{W_{1 1}}$ and the first $m$ columns of $\mtx{W}$, denoted $\mtx{W_{m}}=(\mtx{W_{11}}\mtx{W_{2 1}}^T)^T$, are needed to compute the Nystr\"{o}m approximation:

$$\hat{\mtx{W}} = \mtx{W_{m}}\mtx{W_{1 1}}^{-1}\mtx{W_{m}}^T.$$

The eigenvectors of $\hat{\mtx{W}}$ are then used as an approximation to the eigenvectors of $\mtx{W}$.
Unfortunately, these approximate eigenvectors are not necessarily orthogonal, a property that is necessary for the spectral clustering problem.  However, when $\mtx{W}$ is positive semidefinite, these eigenvectors can be orthogonalized efficiently.  First, we construct

$$
\mtx{Q} = \mtx{W_{11}} + \mtx{W_{m}}^{-\frac{1}{2}}\mtx{W_{2 1}}^{T}\mtx{W_{2 1}}\mtx{W_{m}}^{-\frac{1}{2}}.
$$

Then we compute the eigendecomposition of $\mtx{Q}$ to obtain a matrix $\mtx{U}$ whose columns are equal to the eigenvectors of $\mtx{Q}$ and a diagonal matrix $\mtx{\Lambda}$ with diagonal entries equal to its eigenvalues.  The orthogonalized approximate eigenvectors of $\hat{\mtx{W}}$ are then computed as the columns of

$$
\mtx{V} = \mtx{W_{m}W_{11}}^{-\frac{1}{2}}\mtx{U\Lambda}^{-\frac{1}{2}},
$$

which can be used for clustering.  The Nystr\"om method is thus described as follows in Algorithm~\ref{alg:nystrom}.

\begin{algorithm}{}
	\caption{Nystr\"om Method}\label{alg:nystrom}
\begin{algorithmic}[1]
\Procedure{}{$\mtx{W}$, $m$}\Comment{$n\times n$ similarity matrix $\mtx{W}$, sample size $m$}

	\State Decompose the similarity matrix $\mtx{W}$ as in~\eqref{eq:decomposeW}
	
	\State Compute the approximation $\hat{\mtx{W}} = \mtx{W_{m}}\mtx{W_{1 1}}^{-1}\mtx{W_{m}}^T$
	
	\State Set $\mtx{Q} = \mtx{W_{11}} + \mtx{W_{m}}^{-\frac{1}{2}}\mtx{W_{2 1}}^{T}\mtx{W_{2 1}}\mtx{W_{m}}^{-\frac{1}{2}}$
	
	\State Compute the eigendecomposition of $\mtx{Q}$ to obtain eigenvectors $\mtx{U}$ and eigenvalues $\Lambda$
	
	\State Compute orthogonalization $\mtx{V} = \mtx{W_{m}W_{11}}^{-\frac{1}{2}}\mtx{U\Lambda}^{-\frac{1}{2}}$
	
	\State Use the columns of $\mtx{V}$ as approximate eigenvectors for spectral clustering
	
\EndProcedure
\end{algorithmic}
\end{algorithm}

For this process to work, however, one requires that the similarity matrix $\mtx{W}$ be positive semidefinite.  Therefore, the self-tuning approach given in~\eqref{eq:selftune} can no longer be utilized since it will not necessarily guarantee positive semidefiniteness.
 However, using the fact that choosing $\sigma$ equal to a constant yields a similarity matrix which is positive semidefinite~\cite{cuturi2009positive}, the matrix for this algorithm can be self-tuned by setting
\[
\mtx{W}_{ij}=\exp\left(-{\frac{\|\vct{x_i}/\nu_i-\vct{x}_j/\nu_j||^2}{c}}\right),
\]
where $\nu_i$ is defined in~\eqref{eq:nu}, and $c$ is some fixed constant.

\begin{figure}[ht]
\centering
\subfigure[]{
\includegraphics[width=0.45\textwidth]{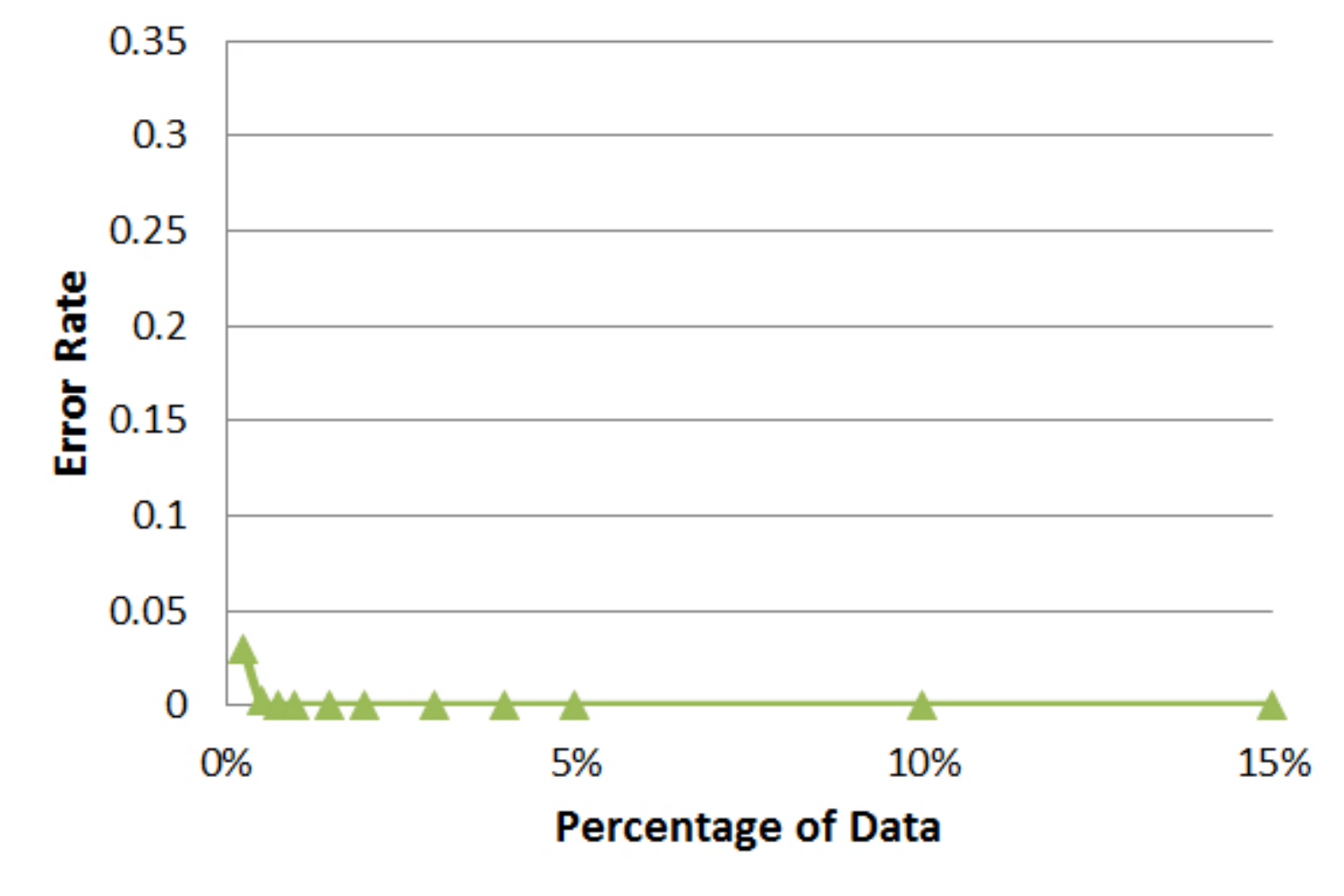}
}
\subfigure[]{
\includegraphics[width=0.45\textwidth]{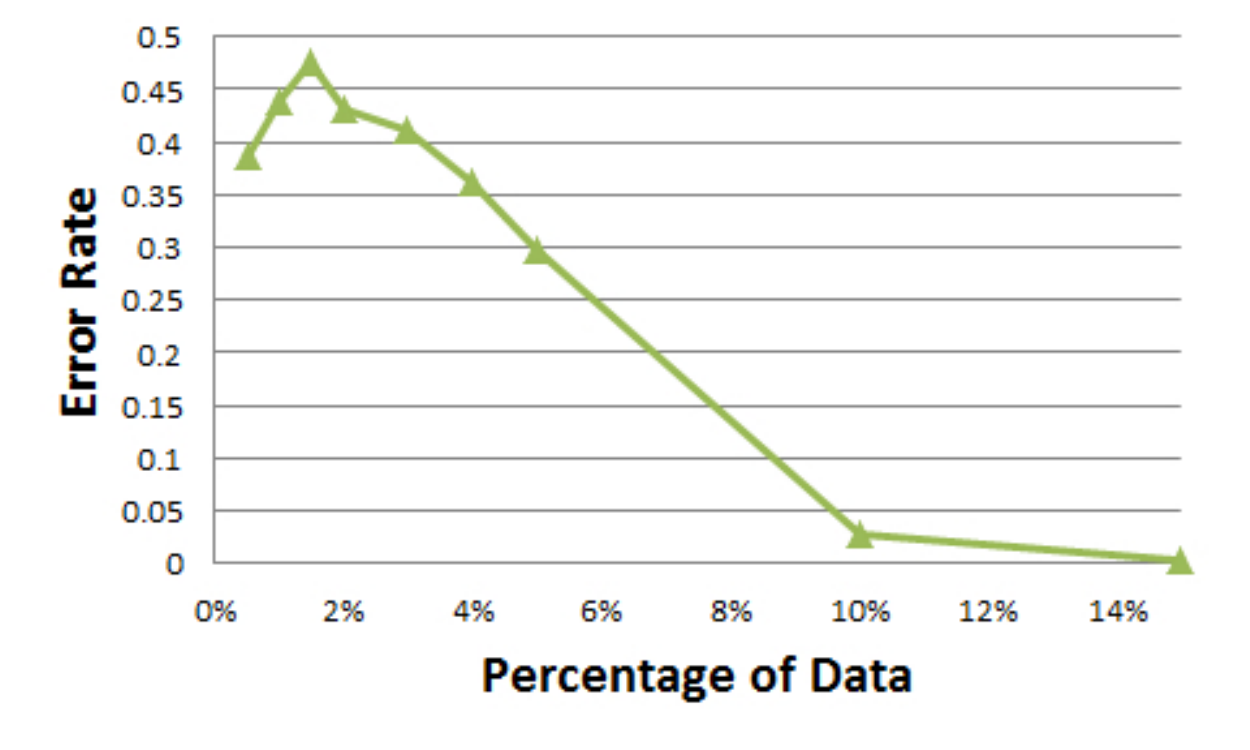}
\label{fig:selftunedsphereerror}
}
\caption[Nystr\"{o}m tuning method comparison]{Manually setting $\sigma$ gives better results than the self-tuning method. An example is shown here for the tangent spheres dataset for (a) $\sigma=1$ and (b) self-tuning. However, different datasets often favor different values of $\sigma$.}\label{fig:sphereerror}
\end{figure}

However, this self-tuning approach yielded worse results empirically than simply manually setting the scaling parameter $\sigma$.  Figure~\ref{fig:sphereerror}
demonstrates the percentage of misclustered data points via the Nystr\"om method with $\sigma = 1$ and the self-tuning approach, for the tangent spheres data (shown in Figure~\ref{fig:SCresults} (e)).

The Nystr\"{o}m method is more efficient than the exact algorithm because it is not necessary to compute eigenvectors of the entire dense similarity matrix.  It has a time complexity of $\bigO(nm^2 + m^3)$, which for $m\ll n$ is significantly less than the $\bigO(n^3)$ runtime of the standard spectral clustering method.

%\notate{Theory? Specific experiments?}

\subsection{Spectral Clustering on a Budget}

The Nystr\"om method relies on a submatrix to approximate the entire similarity matrix $\mtx{W}$.  There, one usually samples blocks or rows/columns at a time.  Alternatively, one can simply sample the \textit{entries} themselves at random.  This is the approach of the \textit{spectral clustering on a budget} method~\cite{shamirspectral}.  The aim is to randomly select $b$ different entries in the matrix (for some budget constraint $b$) and store only those.  The remaining entries are set to zero, enforcing the approximation to be a sparse matrix whose eigenvectors can be computed efficiently.

More specifically, the indices for the entries are chosen uniformly at random and without replacement from $\{(i, j): i < j\}$. A new matrix $\widetilde{\mtx{W}}$ is formed whose entries are given by
\begin{equation}\label{eq:budget}
\widetilde{\mtx{W}}_{ij} = \widetilde{\mtx{W}}_{ji} = \begin{cases}
  \dfrac{2b}{n(n-1)} & \text{if } i = j \\
  \mtx{W}_{ij} & \text{if } (i, j) \text{ is queried} \\
  0 & \text{otherwise}.
\end{cases}
\end{equation}

We thus formulate the spectral clustering on a budget method formally as in Algorithm~\ref{alg:budget}.

\begin{algorithm}{}
	\caption{Spectral clustering on a budget}\label{alg:budget}
\begin{algorithmic}[1]
\Procedure{}{$\mtx{W}$, $m$}\Comment{$n\times n$ similarity matrix $\mtx{W}$, sample size $m$}

	\State Create $\widetilde{\mtx{W}}$ by selecting $m$ entries of $\mtx{W}$ according to~\eqref{eq:budget}
	
	\State Run spectral clustering efficiently using sparsified approximation $\widetilde{\mtx{W}}$
	
\EndProcedure
\end{algorithmic}
\end{algorithm}

It is shown that the perturbation in the eigenvectors via the downsampling in this method can be bounded with high probability.

\begin{theorem}[Spectral clustering on a budget~\cite{shamirspectral}]
Suppose that the budget is $b \leq \frac{1}{4}(n^2 - n)$, and denote by $\vct{v}_2$ and $\widetilde{\vct{v}_2}$ the $2$nd (largest)
eigenvectors of the Laplacian $\mtx{L} = \mtx{D} - \mtx{W}$, and corresponding perturbed Laplacian $\widetilde{\mtx{L}}$, respectively.  Then
$$
\min\left(\|\widetilde{\vct{v}_2} - \vct{v}_2\|_2, \|-\widetilde{\vct{v}_2} - \vct{v}_2\|_2 \right)\lesssim \frac{4}{\lambda_2 - \lambda_3}\left( \frac{n^{5/3}}{b^{2/3}} + \frac{n^{3/2}}{b^{1/2}} \right),
$$
where $\lambda_i$ denotes the $i$th eigenvalue of $\mtx{L}$ and the relation $\lesssim$ ignores lower order logarithmic factors.
\end{theorem}

This result shows that either $\widetilde{\vct{v}_2}$ or $-\widetilde{\vct{v}_2}$ is close to the true eigenvector and can be used for spectral clustering (note that the negative of the eigenvector preserves the same clustering).  This closeness is again controlled by the eigengap of the Laplacian and the budget size $b$.
 If the data are well-clustered, $\mtx{W}$ can be sparsified to have $O(n \log^{3/2} n)$ nonzero entries, and then spectral clustering can be performed in $\bigO(n\log n)$ time.

\section{Numerical Results for Approximation Methods}\label{sec:experiments}

We next describe experimental results for the approximation methods of Section~\ref{sec:approx}.  We run each method using the datasets shown in Figure~\ref{fig:SCresults}.  Each cluster in these sets is clearly defined, and accuracy can thus be easily analyzed.  The aim of these experiments is to compare and analyze the relationship between sample size, runtime, and accuracy for the approximation methods.  We use the convention that a $z\%$ sample size refers to the percentage $z$ of data used in the sample.  For the fast spectral clustering method, this size corresponds to the number of centroids utilized, $k/n$.  The error is reported in terms of the misclustering rate~\eqref{eq:rho}.  Although experiments across different datasets were run on machines of varying specifications, each algorithm for a fixed dataset was run on the same machine to allow for fair comparison.  The algorithms were implemented in Matlab as described in the pseudocode of Section~\ref{sec:approx} and the runtimes were computed via the \textit{cputime} function.  The experiments performed on the interlocked rings, interlocked half rings, and Gaussian strips were run on Intel Core 2 Duo E8500 3.16 GHz machines with 6 MB cache and 16 GB memory.  The concentric spheres and concentric rings experiments were run on Intel Core i7 870 2.93 GHz machines with 4 cores, 8 MB cache and 16 GB memory.  The tangent spheres experiments were run on an Intel Xeon W3520 2.67 GHz machine with 4 cores, 8 MB cache and 16 GB memory.  A constant $\sigma$ value was used for the Nystr\"{o}m method and the spectral clustering on a budget method for the interlocked rings dataset (for the values, see the tables below); otherwise, the self tuning approaches were used. %\notate{Check this with the team.}

\begin{figure}[ht]
\centering
\subfigure[]{
\includegraphics[width=0.45\textwidth]{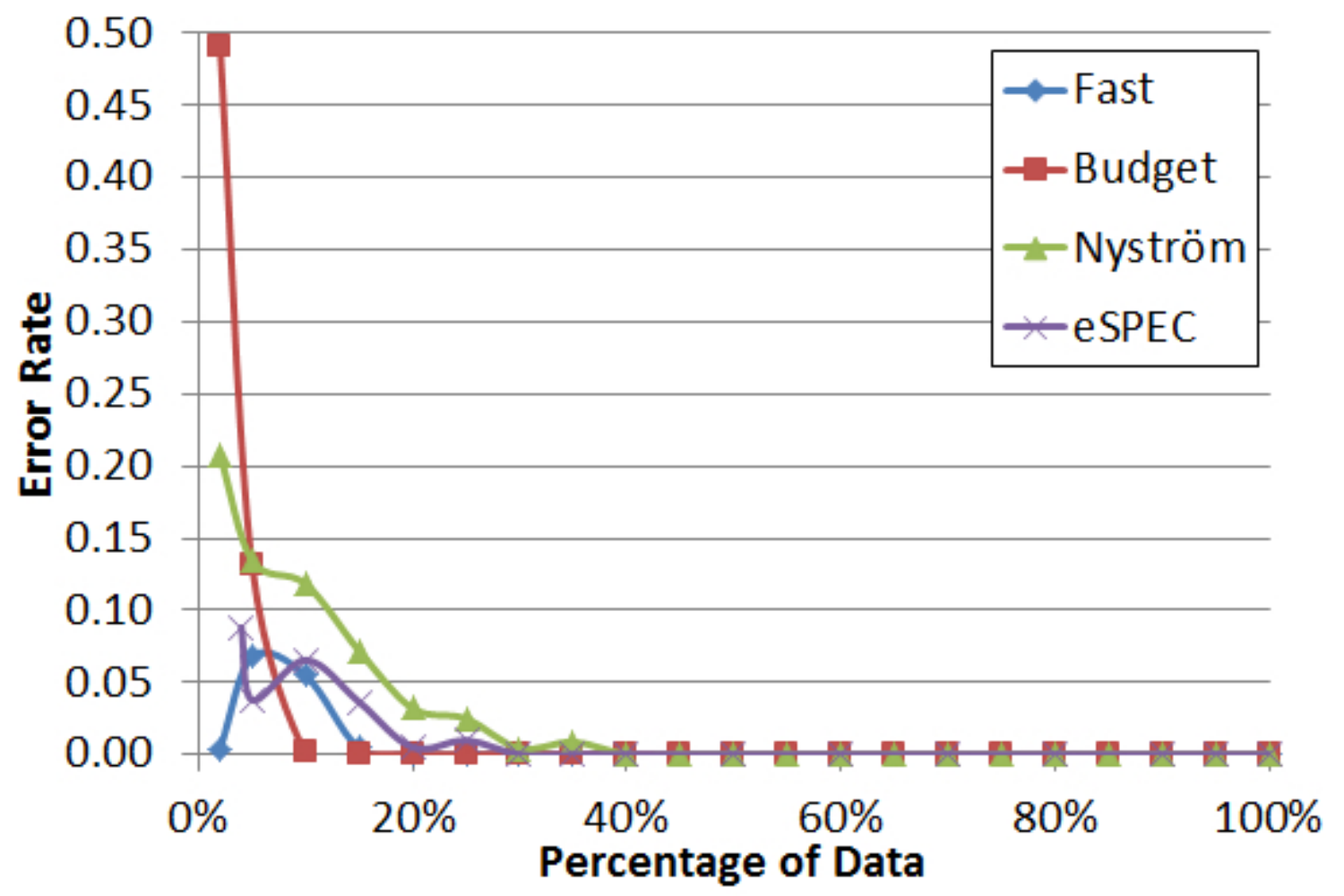}
}
\subfigure[]{
\includegraphics[width=0.45\textwidth]{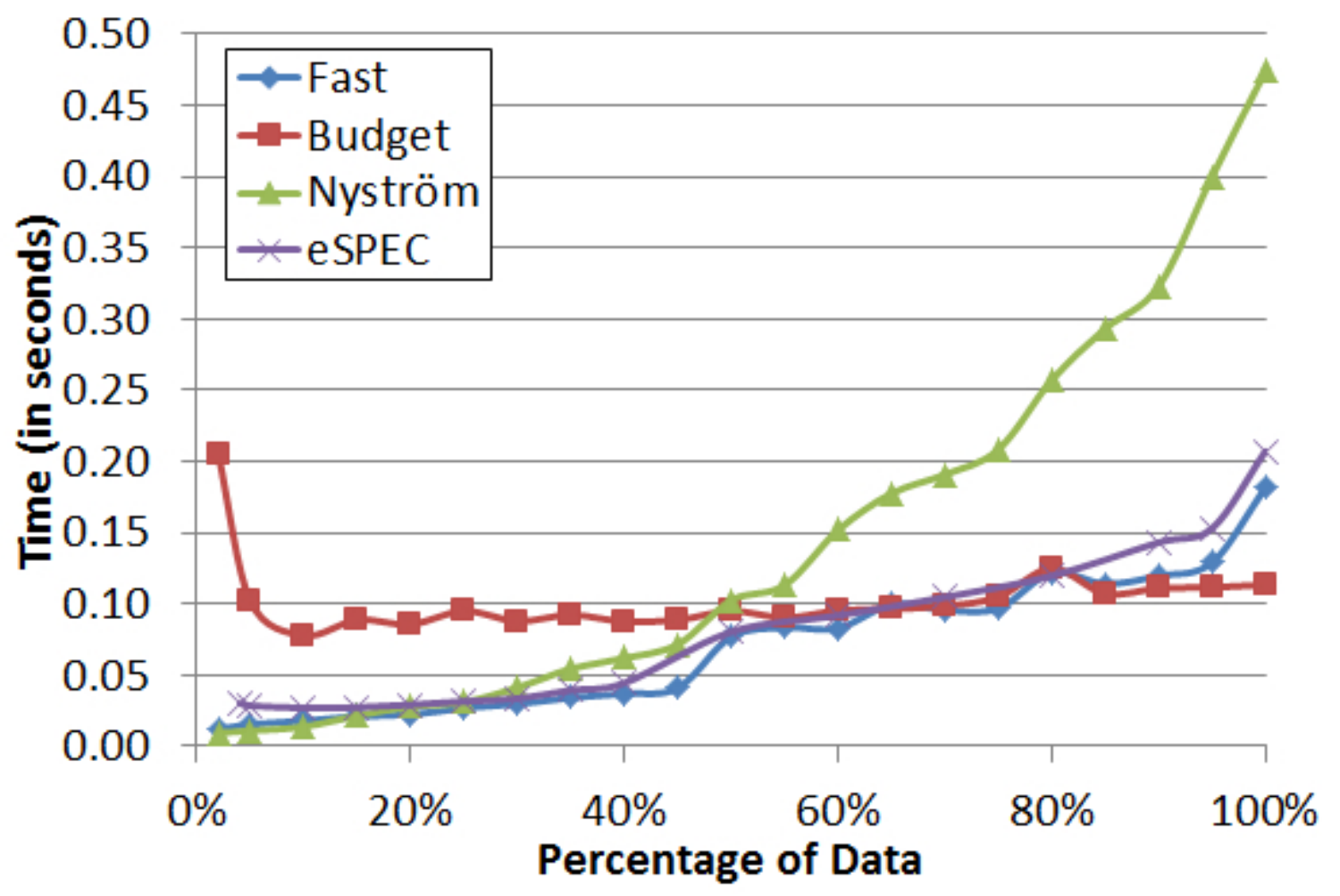}
}
\caption[Gaussian strip dataset Accuracy and Efficiency]{Graphs show the (a) error rate and (b) CPU running time (in seconds) when using different sample sizes for all approximation algorithms. Gaussian strip dataset has $n = 200$ data points and sample sizes range from 0--100\%.}\label{fig:gauss}
\end{figure}

Table~\ref{tab:gauss} and Figure~\ref{fig:gauss} display the results for each algorithm on the Gaussian strips dataset, depicted in Figure~\ref{fig:SCresults} (a).  For this dataset, the error rate and time for the fast spectral clustering method tend to decrease with small enough representative points $k$.  This is most likely because at some point, the $k$-means clustering portion of the algorithm controls how the data clusters.  To get a small error rate for a small enough $k$, the $k$-means clustering must work well with the dataset, in which case spectral clustering is perhaps not necessary (as is most likely the case for a well-separated dataset like the Gaussian strip set).  However, for datasets for which $k$-means does not work well, such as the eye dataset below, we cannot assume that a very small $k$ will have the same accurate results.

As seen in Figure~\ref{fig:gauss}, eSPEC generally performs better than the Nystr\"{o}m method, but fails when we take too small of a sample size.  Spectral clustering on a budget performs worst overall, yielding the highest error rate if too small of a budget is used.  A sufficiently large budget will allow the algorithm to run faster than the original spectral clustering algorithm, but a larger or slightly smaller budget does not significantly change the runtime.  The Nystr\"{o}m method and eSPEC reach a point beyond which an increase in running time fails to produce a commensurate decrease in error rate.  For small sample sizes, time does not change as much, but error rate can increase substantially.  Since the Gaussian data consists of only $n=200$ points, taking a sample as low as $5\%$ might be too small for Nystr\"{o}m and eSPEC to work well.  This is not a problem for fast spectral clustering since the data can be clustered with $k$-means clustering.

\begin{table}[ht]
\small
\begin{center}
\begin{tabular}{|l|l|l|l|l|l|l|l|l|}
\hline
($n=200$) & \multicolumn{2}{c|}{Fast}  & \multicolumn{2}{c|}{Budget}  & \multicolumn{2}{c|}{Nystr\"{o}m ($\sigma = 1$)} & \multicolumn{2}{c|}{eSPEC}\\ \hline
Sample Size & Time & Error & Time & Error & Time & Error & Time & Error\\ \hline
2\% & 0.0122 & 0.0036 & 0.2044 & 0.4914 & 0.0097 & 0.208 &  & \\ \hline
5\% & 0.0156 & 0.068 & 0.1017 & 0.1308 & 0.0112 & 0.1344 & 0.0287 & 0.0379\\ \hline
10\% & 0.0181 & 0.0548 & 0.078 & 0.0015 & 0.014 & 0.1183 & 0.0271 & 0.0656\\ \hline
15\% & 0.0212 & 0.0046 & 0.0889 & 0 & 0.0218 & 0.0719 & 0.0275 & 0.0364\\ \hline
20\% & 0.0225 & 0 & 0.0858 & 0 & 0.0281 & 0.0321 & 0.0293 & 0.005\\ \hline
25\% & 0.0271 & 0 & 0.0952 & 0 & 0.0312 & 0.0249 & 0.0318 & 0.0098\\ \hline
30\% & 0.0300 & 0 & 0.088 & 0 & 0.0415 & 0.0042 & 0.0337 & 0\\ \hline
35\% & 0.0343 & 0 & 0.0924 & 0 & 0.0546 & 0.0088 & 0.0396 & 0\\ \hline
40\% & 0.0371 & 0 & 0.0877 & 0 & 0.0621 & 0 & 0.0446 & 0\\ \hline
50\% & 0.0771 & 0 & 0.0952 & 0 & 0.1026 & 0 & 0.0805 & 0\\ \hline
%60\% & 0.0830 & 0 & 0.0955 & 0 & 0.1519 & 0 & 0.092 & 0\\ \hline
%70\% & 0.0958 & 0 & 0.0986 & 0 & 0.1909 & 0 & 0.1048 & 0\\ \hline
%80\% & 0.1220 & 0 & 0.1251 & 0 & 0.2571 & 0 & 0.1198 & 0\\ \hline
%90\% & 0.1198 & 0 & 0.1111 & 0 & 0.3229 & 0 & 0.1432 & 0\\ \hline
%95\% & 0.1292 & 0 & 0.1117 & 0 & 0.4 & 0 & 0.1532 & 0\\ \hline
%100\% & 0.1816 & 0 & 0.1136 & 0 & 0.4739 & 0 & 0.2072 & 0\\ \hline
\end{tabular}
\end{center}
\caption[Gaussian strips simulation details]{The run time and error rate of each sample size for each approximation algorithm, ran on the Gaussian strip dataset.}\label{tab:gauss}
\end{table}

\begin{figure}[ht]
\centering
\subfigure[]{
\includegraphics[width=0.45\textwidth]{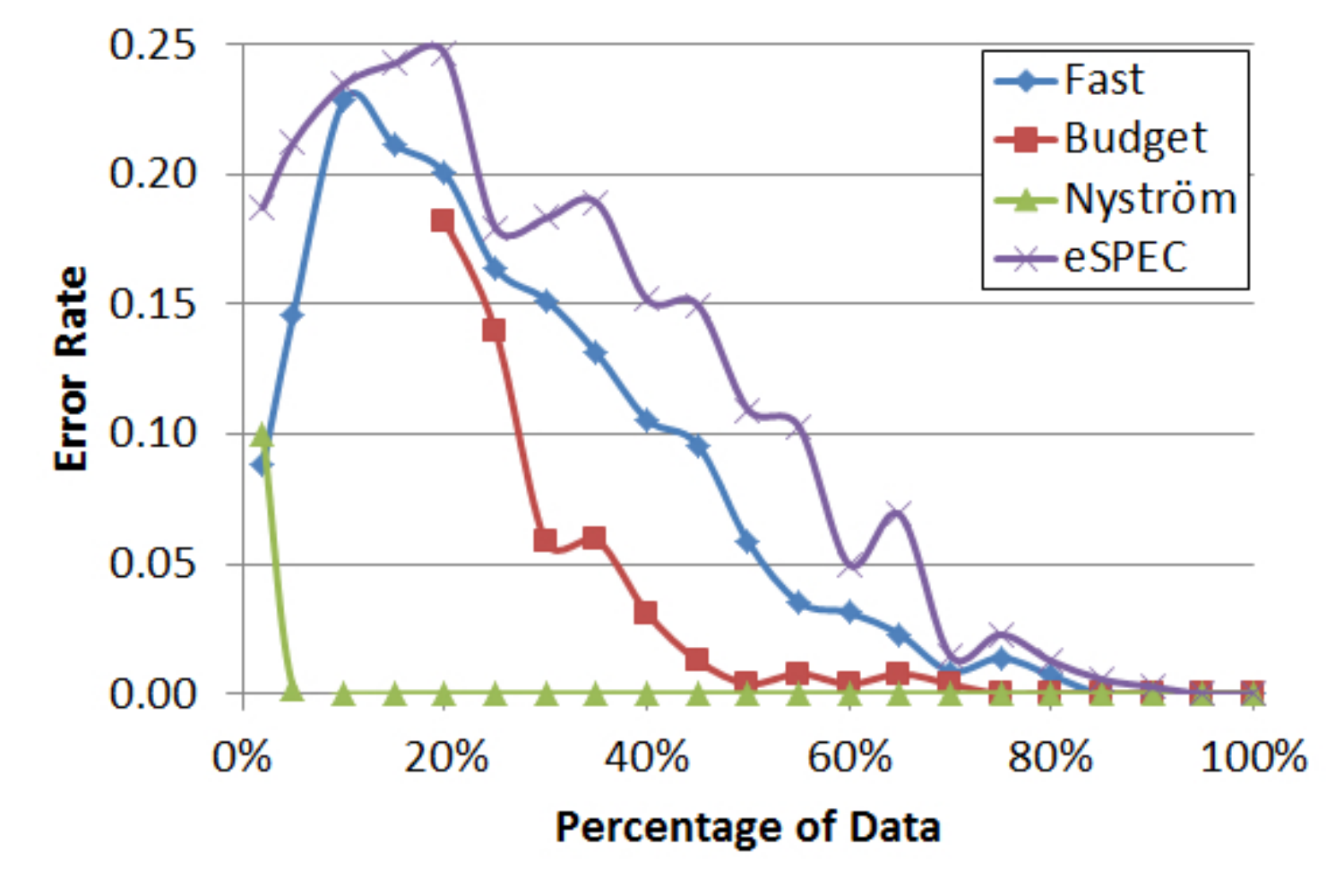}
}
\subfigure[]{
\includegraphics[width=0.45\textwidth]{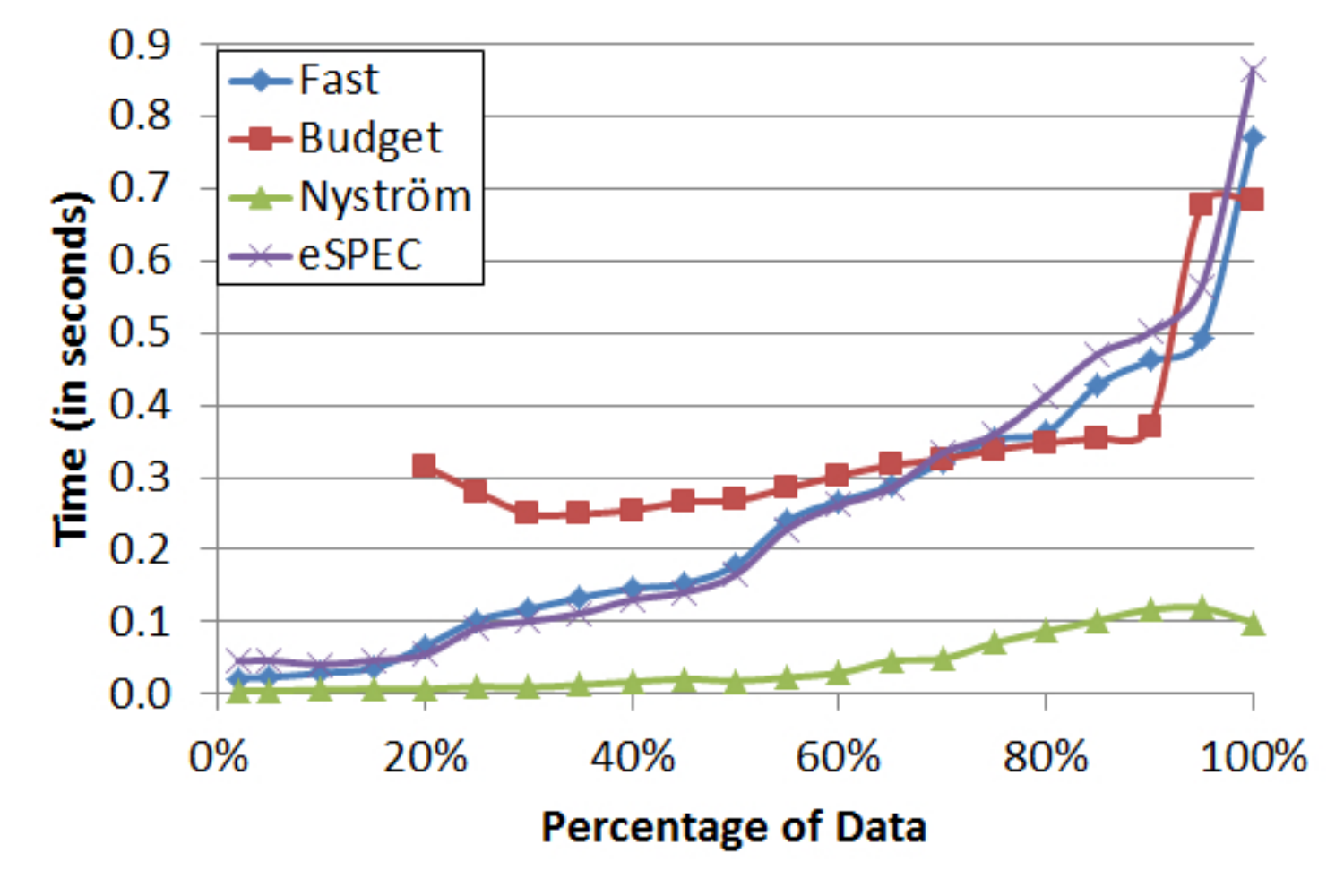}
}
\caption[Toy Problem]{The (a) error rate and (b) CPU running time in seconds when using different sample sizes for all approximation algorithms.  Interlocked half rings dataset has $n = 373$ data points and sample sizes range from 0--100\%.}\label{fig:toy}
\end{figure}

Table~\ref{tab:toy} and Figure~\ref{fig:toy} depict the experimental results for the interlocked half rings dataset, depicted in Figure~\ref{fig:SCresults} (b).
For this dataset, the Nystr\"{o}m method is ideal across the board.  It has the smallest error rate and requires the least amount of time. Fast spectral clustering and eSPEC generally follow the same trend and are very similar in performance.  For small datasets in general, even if $k$-means clustering does not work well with the dataset, fast spectral clustering is effective.  Cells without numerical entries indicate regimes for which the parameters were too small for the algorithm to perform.

\begin{table}[ht]
%sigma and k removed
\small
\begin{center}
\begin{tabular}{|l|l|l|l|l|l|l|l|l|}
\hline
($n=373$) &\multicolumn{2}{c|}{Fast}  & \multicolumn{2}{c|}{Budget}  & \multicolumn{2}{c|}{Nystr\"{o}m ($\sigma = 0.2$)}& \multicolumn{2}{c|}{eSPEC}\\ \hline
Sample Size & Time & Error & Time & Error & Time & Error & Time  & Error\\ \hline
2\% & 0.0215 & 0.0881 & - & - & 0.005 & 0.1 & 0.0452 & 0.1872\\ \hline
5\% & 0.0231 & 0.1461 & - & - & 0.0044 & 0.002 & 0.0462 & 0.2122\\ \hline
%10\% & 0.029 & 0.2286 & - & - & 0.0056 & 0 & 0.0409 & 0.2346\\ \hline
15\% & 0.0356 & 0.2116 & - & - & 0.0069 & 0 & 0.0465 & 0.2426\\ \hline
%20\% & 0.0661 & 0.2003 & 0.3161 & 0.1818 & 0.0066 & 0 & 0.0543 & 0.2466\\ \hline
25\% & 0.1017 & 0.1639 & 0.2805 & 0.1396 & 0.01 & 0 & 0.0908 & 0.1795\\ \hline
%30\% & 0.1173 & 0.1517 & 0.2496 & 0.0583 & 0.009 & 0 & 0.1008 & 0.1832\\ \hline
35\% & 0.1335 & 0.1312 & 0.2493 & 0.0595 & 0.0128 & 0 & 0.1114 & 0.1892\\ \hline
%40\% & 0.1463 & 0.1057 & 0.2546 & 0.0311 & 0.0168 & 0 & 0.1304 & 0.1518\\ \hline
45\% & 0.1529 & 0.0959 & 0.2658 & 0.0129 & 0.0209 & 0 & 0.1407 & 0.1498\\ \hline
%50\% & 0.1785 & 0.0585 & 0.2680 & 0.0039 & 0.0181 & 0 & 0.165 & 0.1094\\ \hline
55\% & 0.2399 & 0.0354 & 0.2855 & 0.0078 & 0.0228 & 0 & 0.2287 & 0.1031\\ \hline
%60\% & 0.2671 & 0.0313 & 0.3026 & 0.0038 & 0.029 & 0 & 0.2615 & 0.0498\\ \hline
65\% & 0.2874 & 0.0229 & 0.3170 & 0.0077 & 0.0456 & 0 & 0.2852 & 0.069\\ \hline
%70\% & 0.3217 & 0.0087 & 0.3254 & 0.0038 & 0.049 & 0 & 0.3342 & 0.0148\\ \hline
75\% & 0.3526 & 0.0139 & 0.3382 & 0 & 0.0708 & 0 & 0.3594 & 0.0229\\ \hline
%80\% & 0.3625 & 0.0071 & 0.3482 & 0 & 0.0867 & 0 & 0.4131 & 0.0126\\ \hline
85\% & 0.4287 & 0 & 0.3547 & 0 & 0.102 & 0 & 0.4711 & 0.0056\\ \hline
%90\% & 0.4621 & 0 & 0.3697 & 0 & 0.117 & 0 & 0.5011 & 0.0028\\ \hline
\end{tabular}
\end{center}
\caption[Toy problem]{The run time and error rate of each sample size for each approximation algorithm, ran on the interlocked half rings dataset.}\label{tab:toy}
\end{table}

\begin{figure}[ht]
\centering
\subfigure[]{
\includegraphics[width=0.45\textwidth]{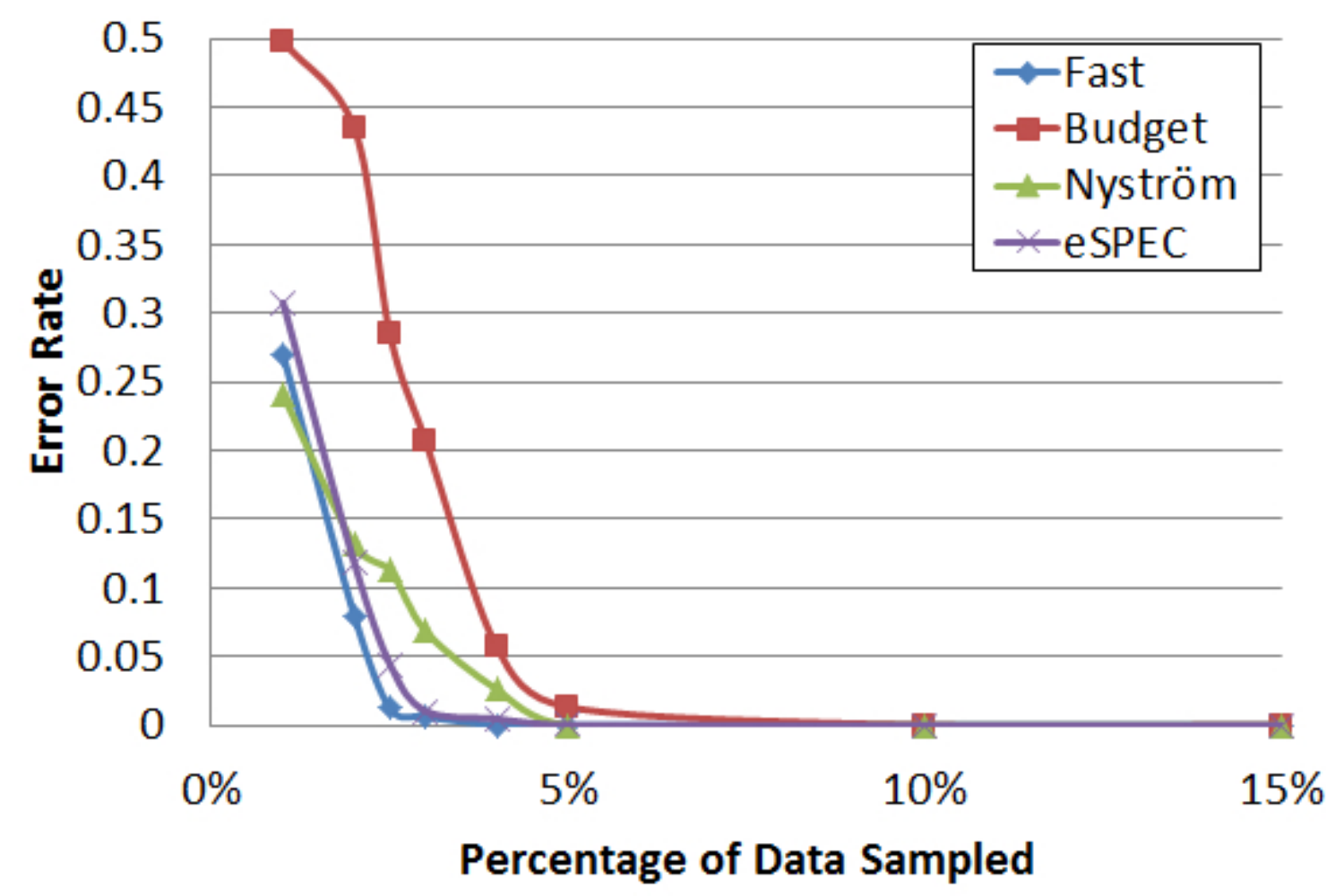}
}
\subfigure[] {
\includegraphics[width=0.45\textwidth]{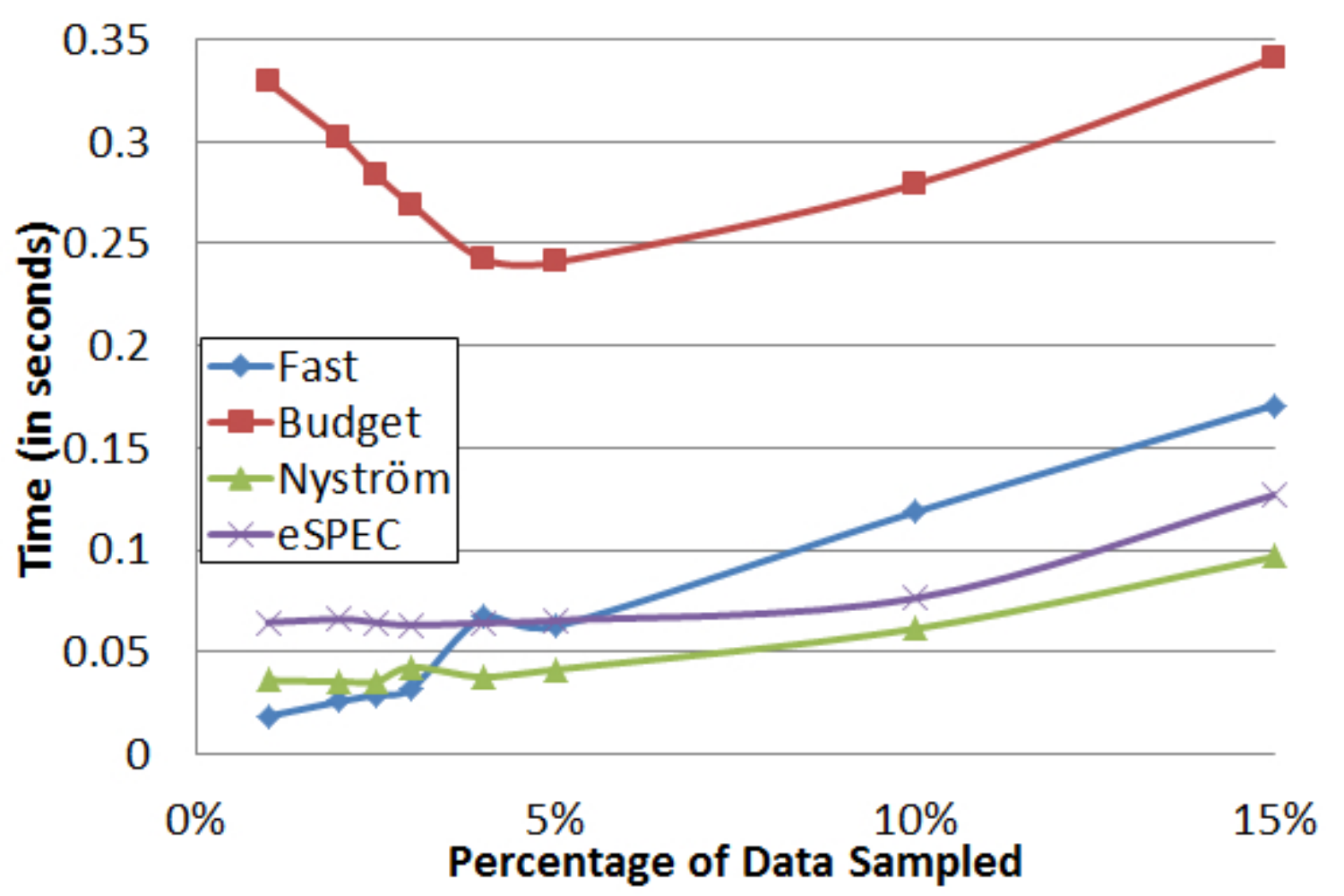}
}
\caption[Eye dataset]{The (a) average error rate and (b) running time when using different sample sizes for all approximation algorithms. Concentric rings dataset has $n = 800$ data points and sample sizes range from 0--15\%.}\label{fig:eye}
\end{figure}

The results obtained from the concentric rings dataset (depicted in Figure~\ref{fig:SCresults} (c)) are displayed in Table~\ref{tab:eye} and Figure~\ref{fig:eye}.  The table shows that spectral clustering on a budget is very inaccurate with this dataset when the sample size is below $3\%$, as it misclusters $1$ in $5$ points.  We see that all methods identify the clusters exactly once $10\%$ of the data is used in the sample.

\begin{table}[ht]
\small
\begin{center}
\begin{tabular}{|l|l|l|l|l|l|l|l|l|}
\hline
($n=800$) & \multicolumn{2}{c|}{Fast}  & \multicolumn{2}{c|}{Budget}  & \multicolumn{2}{c|}{Nystr\"{o}m ($\sigma = 0.5$)}& \multicolumn{2}{c|}{eSPEC}\\ \hline
Sample Size & Time & Error & Time & Error & Time & Error & Time & Error\\ \hline
1.0\% & 0.0187 & 0.2699 & 0.3289 & 0.4979 & 0.0362 & 0.2404 & 0.0646 & 0.3076\\ \hline
2.0\% & 0.0259 & 0.0792 & 0.3017 & 0.4341 & 0.0356 & 0.1311 & 0.0661 & 0.1176\\ \hline
2.5\% & 0.0291 & 0.012 & 0.2839 & 0.2841 & 0.0356 & 0.1127 & 0.0646 & 0.0441\\ \hline
3\% & 0.0324 & 0.0062 & 0.2689 & 0.2061 & 0.0427 & 0.0682 & 0.0633 & 0.0102\\ \hline
4\% & 0.0674 & 0 & 0.2424 & 0.0564 & 0.0378 & 0.0258 & 0.0643 & 0.0046\\ \hline
5\% & 0.0627 & 0 & 0.2409 & 0.0131 & 0.0415 & 0 & 0.0655 & 0.0001\\ \hline
10\% & 0.1186 & 0 & 0.2792 & 0 & 0.0615 & 0 & 0.0764 & 0\\ \hline
15\% & 0.171 & 0 & 0.341 & 0 & 0.0967 & 0 & 0.127 & 0\\ \hline
\end{tabular}
\end{center}
\caption[]{The (a) run time and (b) error rate of each sample size for each approximation algorithm, ran on the concentric rings dataset.}\label{tab:eye}

\end{table}

\begin{figure}[ht]
\centering
\subfigure[]{
\includegraphics[width=0.45\textwidth]{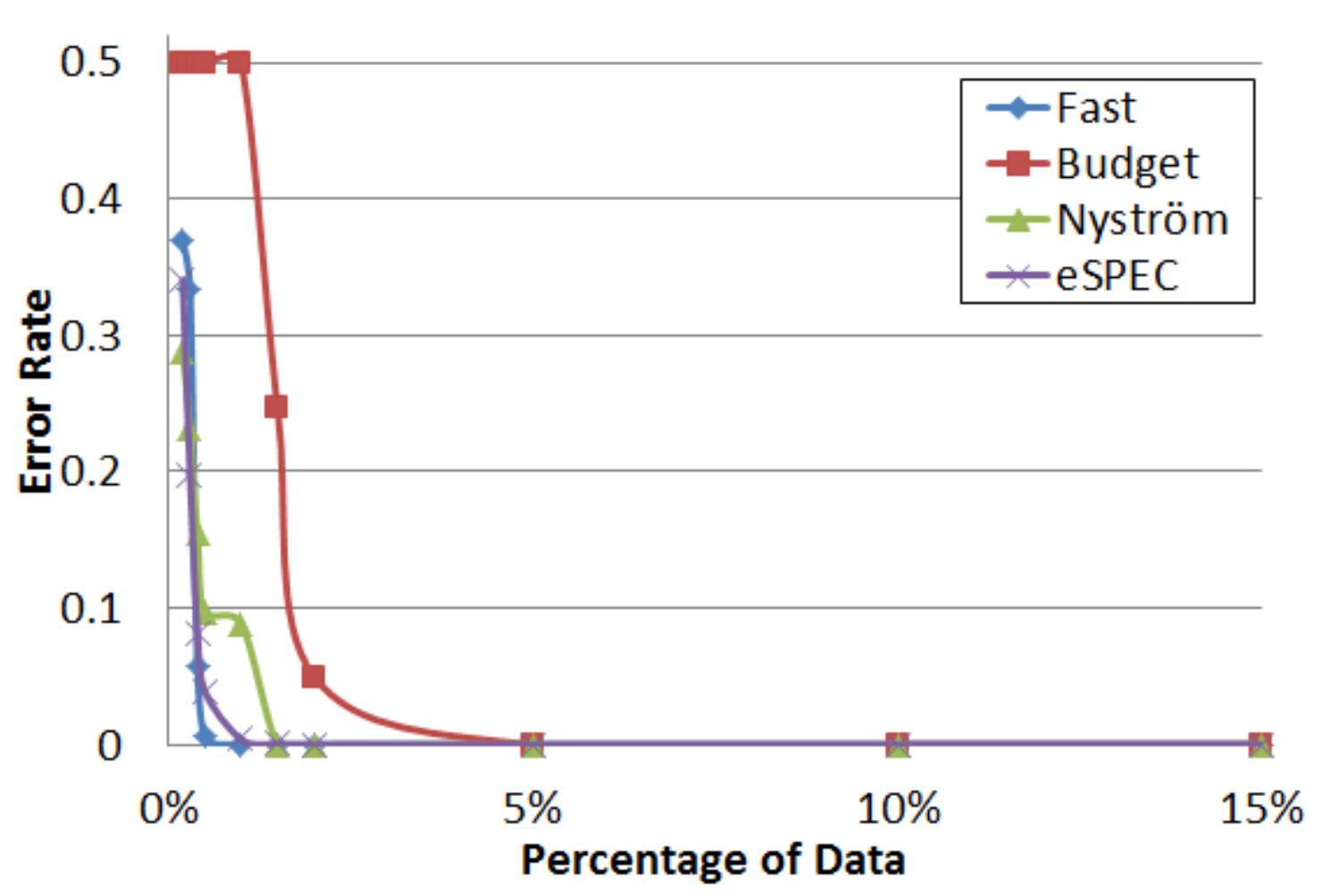}
}
\subfigure[]{
\includegraphics[width=0.45\textwidth]{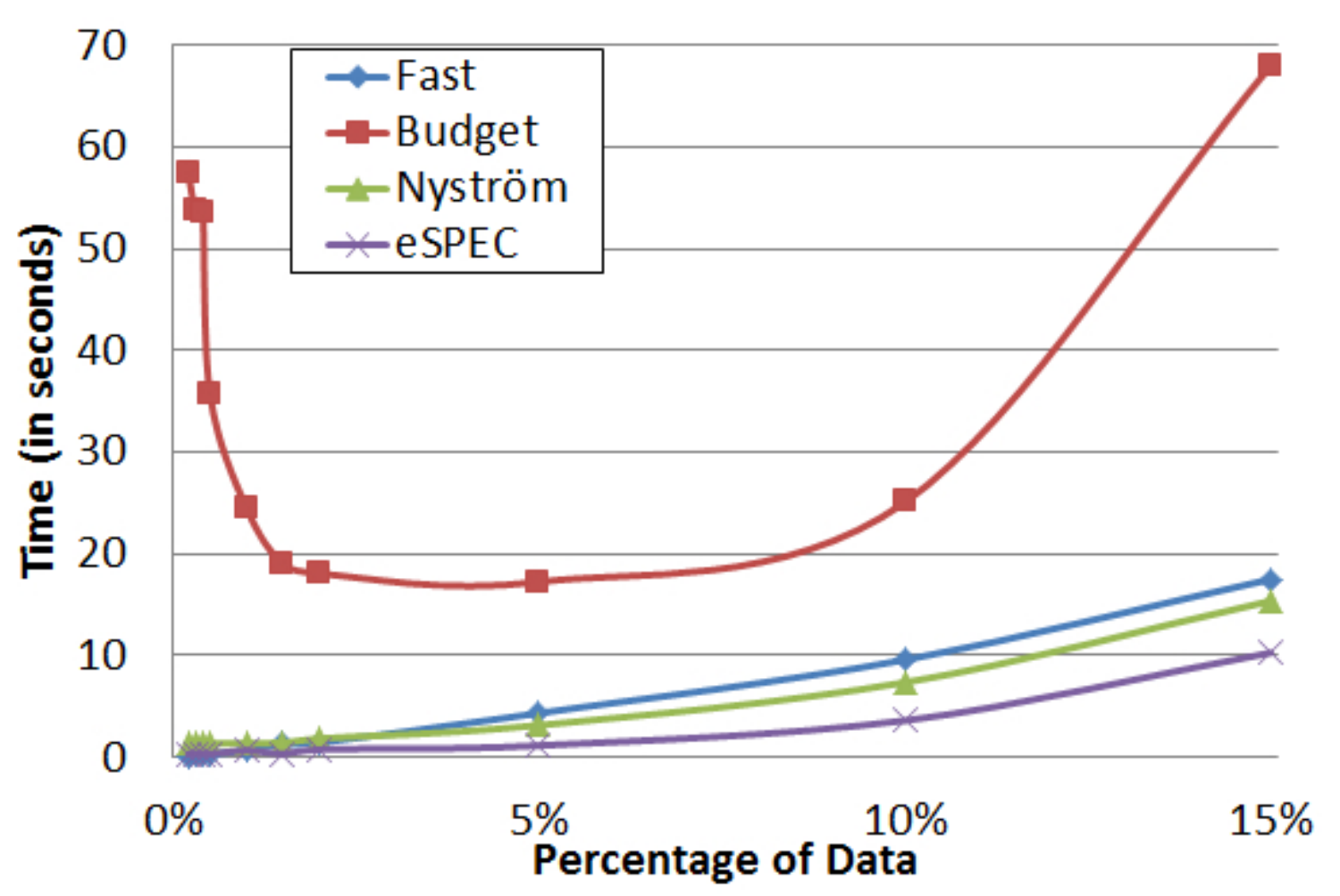}
}
\caption[Coconut]{The (a) average error rate and (b) running time when using different sample sizes for all approximation algorithms. Concentric spheres dataset has $n = 5{,}000$ data points and sample sizes range from 0--15\%.}\label{fig:coco}
\end{figure}

Table \ref{tab:coco} and Figure~\ref{fig:coco} contain the results of the experiments on the concentric spheres dataset, depicted in Figure~\ref{fig:SCresults} (d).  As the three dimensional analog of the concentric rings, it is not surprising we see similar results.  It is to be noted that all algorithms cluster perfectly when the sample size is $5\%$ or larger.

\begin{table}[ht]
\small
\begin{center}
\begin{tabular}{|l|l|l|l|l|l|l|l|l|}
\hline
($n=5,000$) & \multicolumn{2}{c|}{Fast}  & \multicolumn{2}{c|}{Budget}  & \multicolumn{2}{c|}{Nystr\"{o}m ($\sigma = 1$)} & \multicolumn{2}{c|}{eSPEC}\\ \hline
Sample Size & Time & Error & Time & Error & Time & Error & Time & Error\\ \hline
0.2\% & 0.1292 & 0.3688 & 57.4396 & 0.4998 & 1.3934 & 0.2884 & 0.356 & 0.3405\\  \hline
0.3\% & 0.2296 & 0.3334 & 53.7683 & 0.4996 & 1.3672 & 0.2309 & 0.3588 & 0.1967\\  \hline
0.4\% & 0.3404 & 0.0569 & 53.6019 & 0.4997 & 1.3837 & 0.1539 & 0.3622 & 0.0812\\  \hline
0.5\% & 0.3644 & 0.0064 & 35.6188 & 0.4995 & 1.3937 & 0.0976 & 0.3469 & 0.0394\\  \hline
1.0\% & 0.7033 & 0 & 24.5065 & 0.4993 & 1.4212 & 0.089 & 0.6964 & 0.0041\\  \hline
1.5\% & 1.3407 & 0 & 18.9463 & 0.246 & 1.4546 & 0 & 0.3716 & 0.0006\\  \hline
2.0\% & 1.487 & 0 & 18.0827 & 0.0498 & 1.87 & 0 & 0.805 & 0\\  \hline
5.0\% & 4.3699 & 0 & 17.2624 & 0 & 3.1868 & 0 & 1.1784 & 0\\  \hline
10.0\% & 9.6112 & 0 & 25.099 & 0 & 7.3857 & 0 & 3.6448 & 0\\  \hline
15.0\% & 17.528 & 0 & 67.9837 & 0 & 15.3998 & 0 & 10.3045 & 0\\  \hline
\end{tabular}
\end{center}
\caption[Coconut]{The run time and error rate of each sample size for each approximation algorithm, ran on the concentric spheres dataset.}\label{tab:coco}
\end{table}

\begin{figure}[ht]
\centering
\subfigure[]{
\includegraphics[width=0.45\textwidth]{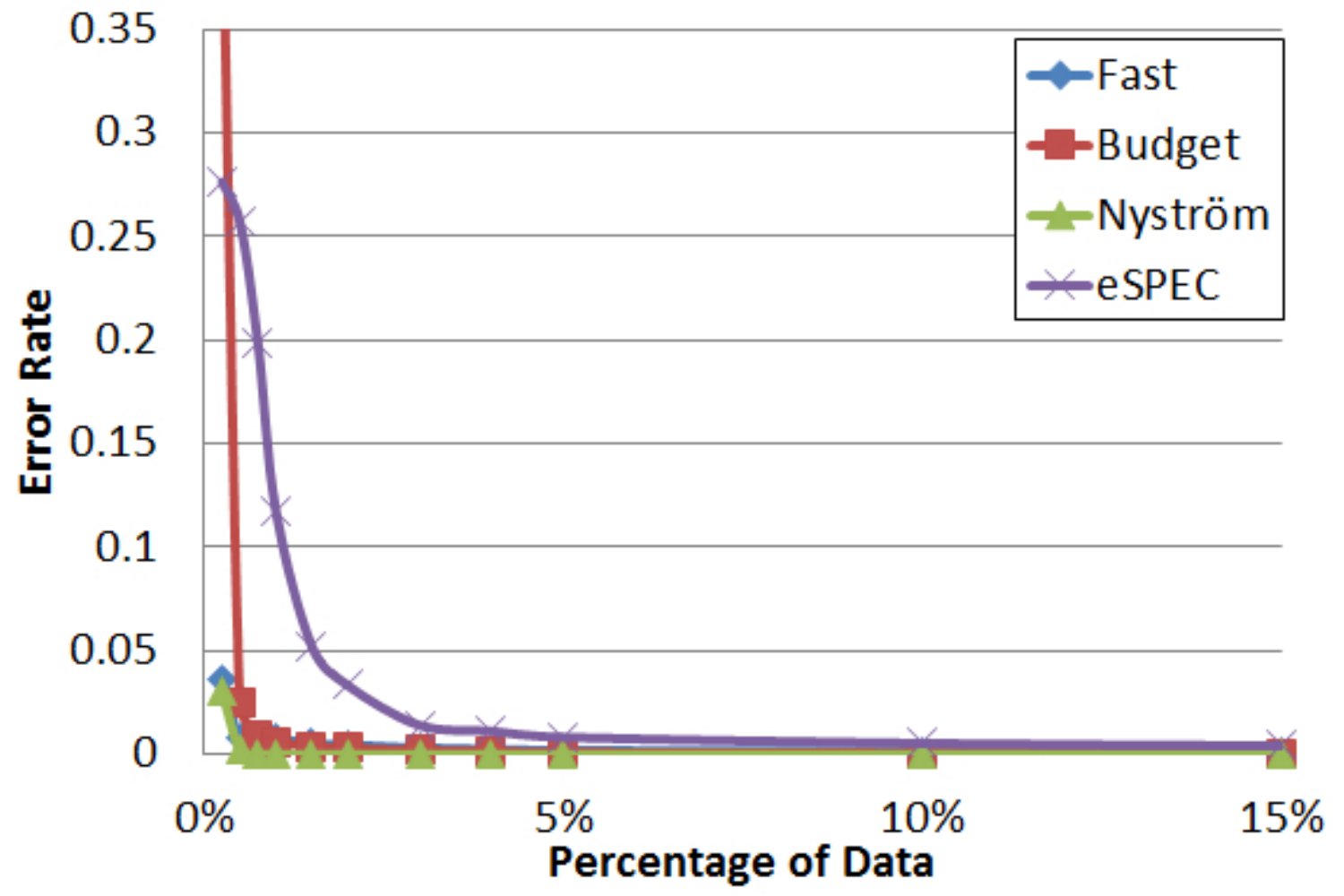}
}
\subfigure[]{
\includegraphics[width=0.45\textwidth]{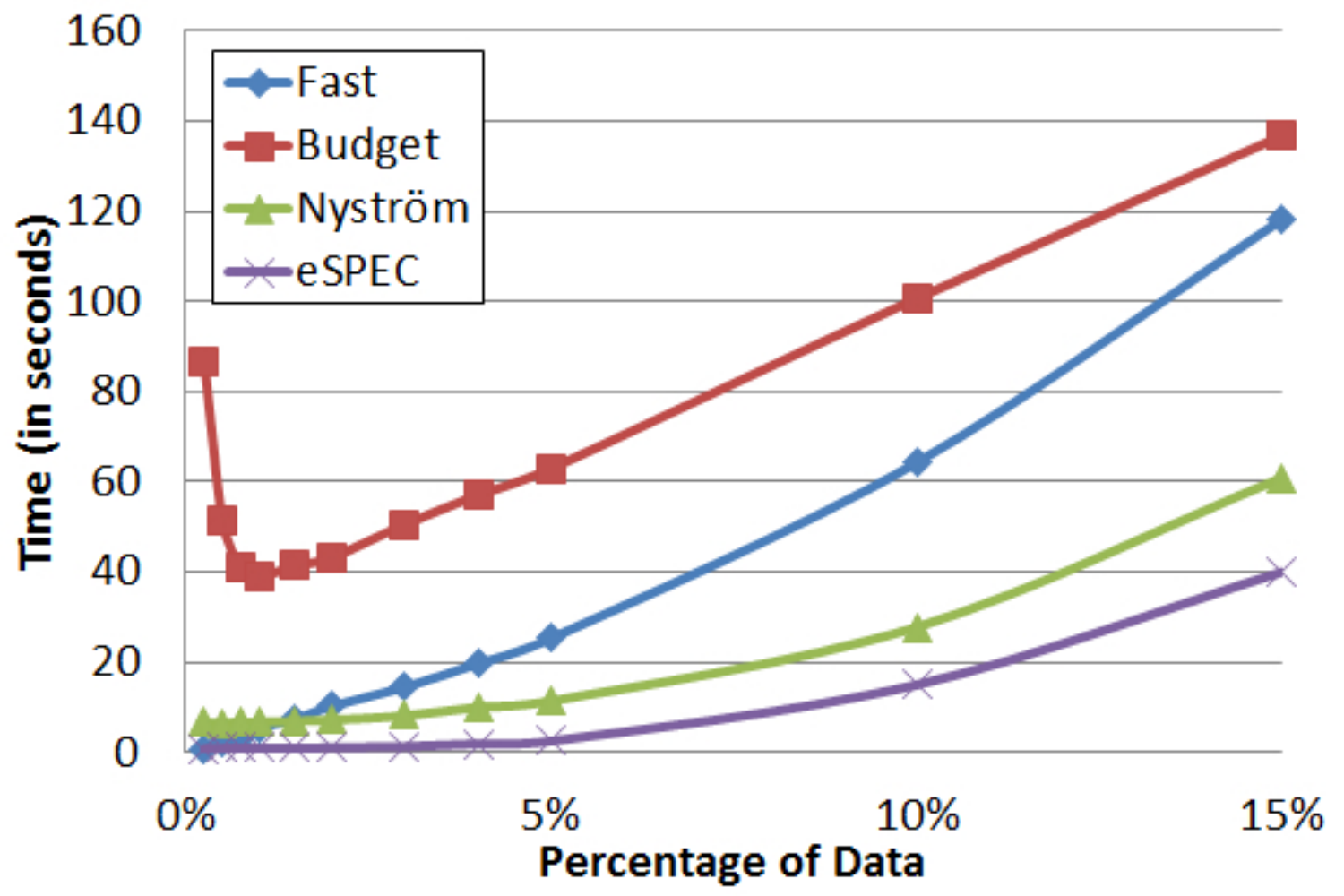}
}
\caption[Spheres]{The (a) average error rate and (b) CPU running time in seconds when using different sample sizes for all approximation algorithms.  Tangent spheres dataset has $n = 10{,}000$ data points and sample sizes range from 0--15\%.}\label{fig:tan}
\end{figure}

Table~\ref{tab:tan} and Figure~\ref{fig:tan} depict the results of the algorithms run on the tangent spheres dataset, depicted in Figure~\ref{fig:SCresults} (e).  This dataset gives us a prime example of how approximate spectral clustering algorithms are ideal for large, structured data.  The exact algorithm takes $6,365.4$ seconds, or almost $2$ hours, to give results, when nearly identical results can be given in seconds using approximations.  When just $4.25\%$ of the data is sampled, all of the algorithms perform with error less than $0.01$, and all under a minute.

\begin{table}[ht]
\small
\begin{center}
\begin{tabular}{|l|l|l|l|l|l|l|l|l|}
\hline
($n=10{,}000$) & \multicolumn{2}{c|}{Fast}  & \multicolumn{2}{c|}{Budget}  & \multicolumn{2}{c|}{Nystr\"{o}m ($\sigma = 1$)} & \multicolumn{2}{c|}{eSPEC}\\ \hline
Sample Size & Time & Error & Time & Error & Time & Error & Time & Error\\ \hline
0.25\% & 0.8022 & 0.036 & 86.5802 & 0.4268 & 6.647514612 & 0.030166 & 0.8833 & 0.2763\\ \hline
0.50\% & 1.9678 & 0.0075 & 51.4463 & 0.0247 & 6.592 & 0.0021 & 0.9316 & 0.2571\\ \hline
0.75\% & 3.428 & 0.0097 & 41.0183 & 0.0087 & 6.6759 & 0.00037 & 0.9382 & 0.1987\\ \hline
1\% & 5.5 & 0.0081 & 39.0598 & 0.0061 & 6.7492 & 0.000118 & 0.9734 & 0.1177\\ \hline
1.5\% & 7.4556 & 0.0052 & 41.432 & 0.0039 & 6.9648 & 0.000078 & 1.0056 & 0.0524\\ \hline
2\% & 10.3042 & 0.0047 & 43.2666 & 0.003 & 7.2534 & 0.000072 & 1.1082 & 0.0334\\ \hline
3\% & 14.5371 & 0.0032 & 50.6176 & 0.002 & 8.1726 & 0.00009 & 1.3419 & 0.0141\\ \hline
4\% & 19.7435 & 0.0025 & 57.1082 & 0.0015 & 10.0714 & 0.0001 & 1.9394 & 0.0112\\ \hline
5\% & 25.3318 & 0.0019 & 63.0219 & 0.0011 & 11.4333 & 0.0001 & 2.5556 & 0.0083\\ \hline
10\% & 64.279 & 0.0016 & 100.6472 & 0.000656 & 27.7863 & 0.0001 & 15.0004 & 0.0054\\ \hline
15\% & 118.1536 & 0.0013 & 136.7211 & 0.000466 & 60.9053 & 0.0001 & 40.0202 & 0.0042\\
\hline
\end{tabular}
\caption[Spheres]{The run time and error rate of each sample size for each approximation algorithm, ran on the tangent spheres dataset.}
\label{tab:tan}
\end{center}
\end{table}

\begin{figure}[ht]
\centering
\subfigure[]{
\includegraphics[width=0.45\textwidth]{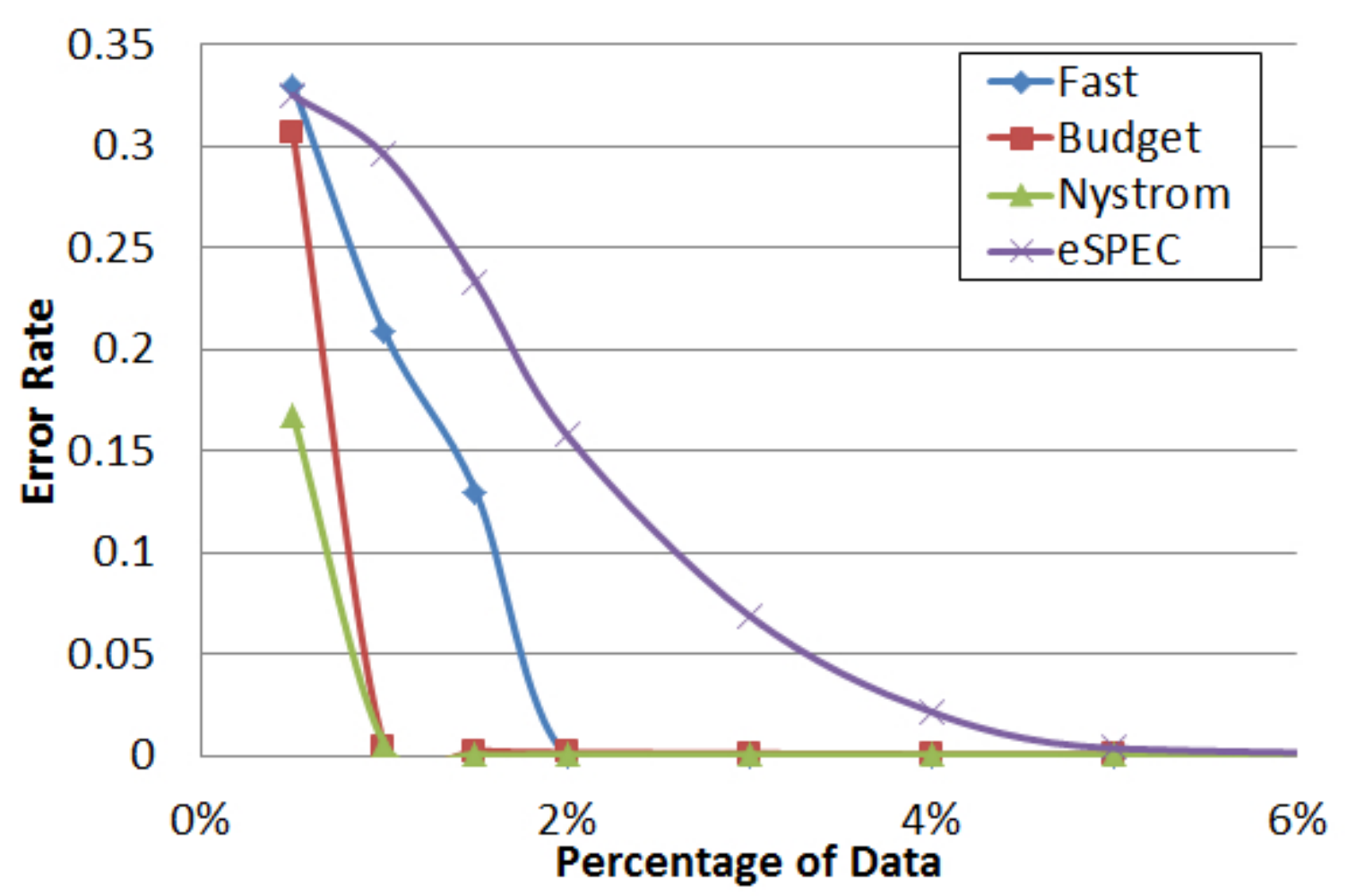}
}
\subfigure[]{
\includegraphics[width=0.45\textwidth]{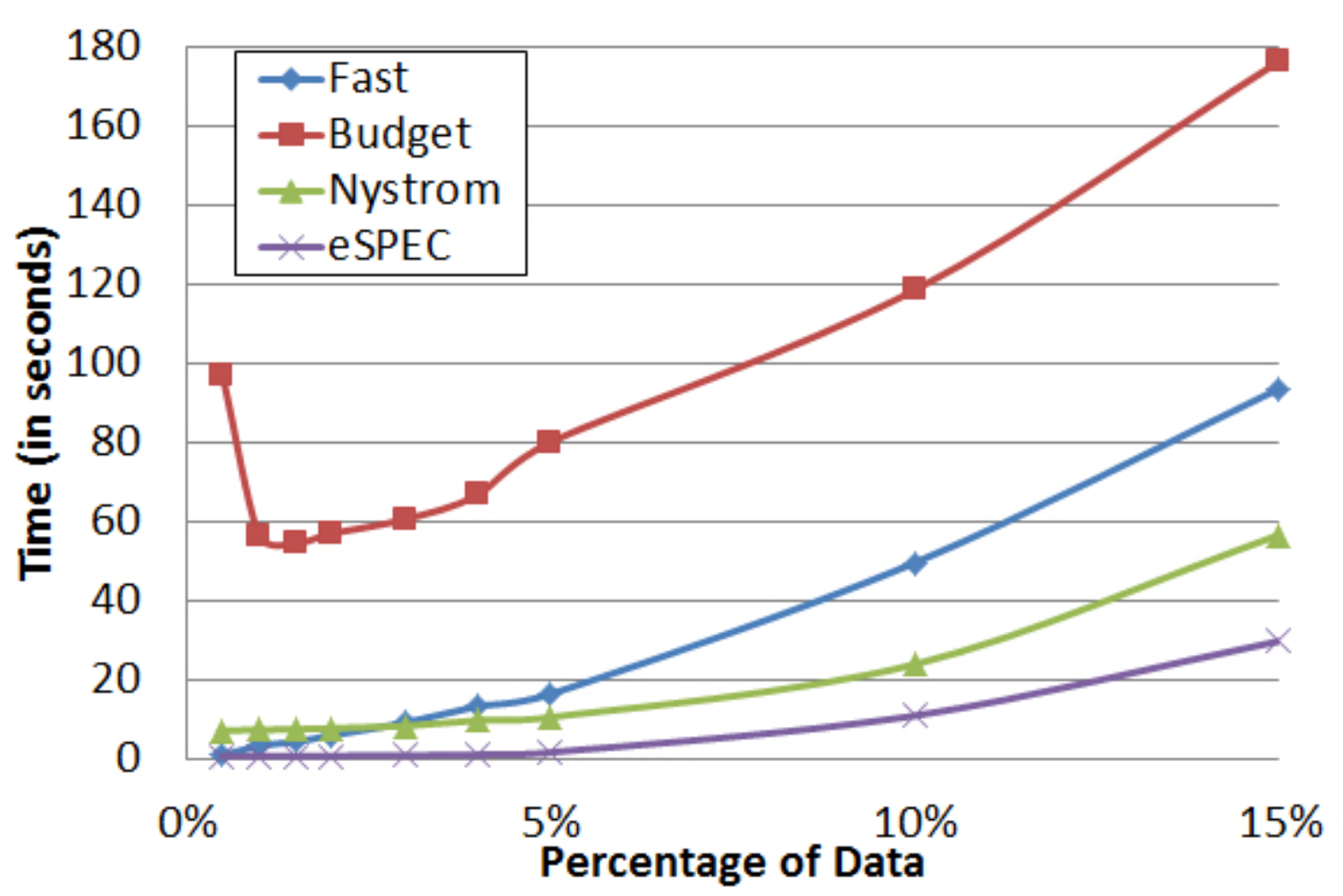}
}
\caption[Spheres]{The (a) average error rate and (b) CPU running time in seconds when using different sample sizes for all approximation algorithms.  Interlocked rings dataset has $n = 10{,}000$ data points and sample sizes range from 0--15\%.}\label{fig:donuts}
\end{figure}

Finally, Table~\ref{tab:donuts} and Figure~\ref{fig:donuts} demonstrate the results of the algorithms run on the interlocked rings dataset of Figure~\ref{fig:SCresults} (f).  We again see similar results to the tangent spheres dataset, most likely because the structure of the clusters are similar.

\begin{table}[ht]
\small
\begin{center}
\begin{tabular}{|l|l|l|l|l|l|l|l|l|}
\hline
($n=10{,}000$) & \multicolumn{2}{c|}{Fast}  & \multicolumn{2}{c|}{Budget ($\sigma = 0.5$)}  & \multicolumn{2}{c|}{Nystr\"{o}m ($\sigma = 0.5$)} & \multicolumn{2}{c|}{eSPEC}\\ \hline
Sample Size & Time & Error & Time & Error & Time & Error & Time & Error\\ \hline
$0.5\%$  & 1.136 & 0.3298  & 96.986 & 0.3064  & 7.429 & 0.1676  & 0.8998 & 0.3252\\ \hline
$1 \%$ & 3.554 & 0.209  & 56.876 & 0.0048   & 7.575 & 0.0059  & 0.9284 & 0.2958\\ \hline
$1.5 \%$ & 4.538 & 0.1302  & 54.718 & 0.0027 & 7.759 & 0.0011  & 0.9466 & 0.2335\\ \hline
$2\%$  & 6.21 & 0  & 57.23 & 0.0018 &  7.966 & 0.0009  & 0.9962 & 0.1579\\ \hline
$3\%$  & 9.41 & 0  & 60.79 & 0.0014 & 8.595 & 0.001  & 1.1128 & 0.0691\\ \hline
$4\%$  & 13.552 & 0  & 66.9 & 0.001 & 9.947 & 0.0009  & 1.3298 & 0.0217\\ \hline
$5\%$  & 16.64 & 0  & 80.278 & 0.0008 & 10.778 & 0.001  & 1.9226 & 0.0039\\ \hline
$10\%$  & 49.814 & 0  & 118.72 & 0.0008 & 24.164 & 0.001  & 11.185 & 0\\ \hline
$15\%$  & 93.782 & 0  & 176.44 & 0.0007 & 56.83 & 0.001  & 29.976 & 0\\ \hline
\end{tabular}
\caption{The run time and error rate of each sample size for each approximation algorithm, ran on the interlocked rings dataset. }\label{tab:donuts}
\end{center}
\end{table}

\section{Case Study: The Attrition Problem}\label{sec:attrition}

Clustering methods have a wide range of applications, ranging from image segmentation to social network identification.  An application we focus on here is the attrition problem.  In this setting, the data objects correspond to employees of a particular company, and each data vector contains a list of employee attributes.  For example, age, salary, years at the company, number of children, age of children, etc. may all be relevant attributes.  From this data, one wishes to distinguish a cluster of employees who are likely to leave the company from the other cluster of employees are likely to stay with the company.  Such a classification allows companies to invest resources appropriately in an effort to maintain desired employees, saving significant expense and training time.  We analyze here how this problem can be solved using spectral clustering methods.  Because the number $n$ of employees may be very large, and the number of attributes collected about the employees may also be very large, approximation methods are crucial to solve this problem efficiently.  In contrast to the examples of Section~\ref{sec:experiments}, datasets in this setting are not only high dimensional, but accuracy is often difficult to quantify since there may no longer be a notion of ``correct'' clusters.  To overcome this last challenge, we utilize historical data about teacher attrition which will allow us to properly identify the appropriate clustering.

Each year, the National Center for Education Statistics sends out a follow-up survey to teachers originally selected for the Teacher Questionnaire in a Schools and Staffing Survey.  $4{,}528$ teachers were given one of two surveys according to their employment status.  Teachers were classified as either \textit{stayers}, \textit{movers}, or \textit{leavers}.  Stayers are teachers who stayed at their current position, movers are teachers who continued teaching, but transferred schools, and leavers are teachers who left the position entirely.  Leavers took the former teacher questionnaire while stayers and movers took the current teacher questionnaire.  Both surveys contained different sets of questions; the dataset used in our experiments is made up of common questions in both surveys from the 1994--1995 school year.  The attributes include household income (broken into intervals), marital status (coded as $0/1/2$ for never married / married / separated), number of dependent children, age of youngest child, and dissatisfaction ratings.
Because many teachers had the same responses in the six variables, a dummy variable was added so that the algorithm would recognize that the teachers are different people.  The dummy variable was drawn uniformly between $0$ and $1$.

Spectral clustering of the entire dataset yielded a small cluster on 197 teachers who were never married and did not have children or other dependents.  All of their household incomes ranged from $\$60{,}000$ to $\$74{,}999$ and none of the teachers expressed dissatisfaction in the survey.  The number of stayers, movers, and leavers are summarized in Table \ref{TABLE:teacher_nochildren}.
%\begin{table}[ht]
%\begin{center}
%\begin{tabular}{|l|cc|}
%\hline
%Status & Cluster 1 & Cluster 2 \\ \hline
%Stayer & 92 & 1,666\\
%Mover & 24 & 1,016\\
%Leaver & 81 & 1,649\\ \hline
%Total & 197 & 4,331\\ \hline
%\end{tabular}
%\end{center}
%\caption[Teacher dataset, all]{Resulting clusters of the entire teacher dataset.}\label{TABLE:teacher_all}
%\end{table}

Drawing conclusions about the likelihood of attrition for a group of teachers depends on the classification of movers.  From a school's point of view, a mover is an attritor, but from the state's point of view, a mover is still a teacher.  In Table \ref{TABLE:teacher_nochildren}, if movers are considered leavers, then the proportion of attritors in Cluster 1 is significantly different from the proportion of attritors in Cluster 2.  The results yield a one-tailed $p$-value of $0.01$ where teachers with the same characteristics as Cluster 1 are less likely to resign.  However, if movers are considered stayers, then the proportion of attritors in Cluster 1 is not significantly different from the proportion of attritors in Cluster 2.  In this case, the one-tailed $p$-value is $0.1950$ where teachers with the same characteristics as Cluster 1 are more likely to resign.  Because conclusions were affected by the classification of teachers who moved, these teachers were not included in most experiments.  However, one can easily use the applications of spectral clustering to cases where teachers are considered either one or the other.

While the age of a teacher's youngest child can provide valuable information about the teacher's age itself, setting different values for this variable for teachers without children gave varying results.  For example, if the youngest child's age is set to $50$, a large distance away from the maximum value of $38$, spectral clustering tends to group teachers with older children together with childless teachers.  If the youngest child's age is set to $-1$, a closer but impossible age, spectral clustering tends to group teachers with newborns together with childless teachers.  In the following experiments, we used $-1$ as the age for teachers without children, though the technicalities of this variable bring us to question if interaction terms ought to be considered when working with data of this nature.

Spectral clustering was applied on subgroups of teachers, such as teachers without children or unmarried teachers.  Most experiments yielded clusters with a mix of teachers who stayed and teachers who left.  If teachers who moved were included, they would generally be mixed in both clusters as well.  An example of this can be seen in Table \ref{TABLE:teacher_nochildren} where we applied spectral clustering to teachers with children; movers were removed.  Although both clusters contained a mix of stayers and leavers, in a two-proportion $Z$-test, we obtain a one-tailed $p$-value of about 0.0023, providing evidence that Cluster 2 has a greater proportion of teachers who will quit.  We conclude that teachers with similar characteristics as those in Cluster 2 are more likely to quit.

%\begin{table}[ht]
%\begin{center}
%\begin{tabular}{|l|cc|}
%\hline
%Status & Cluster 1 & Cluster 2 \\ \hline
%Stayer & 100 & 870 \\
%Leaver & 48 & 699 \\ \hline
%Total & 148 & 1,569 \\ \hline
%\end{tabular}
%\end{center}
%\caption[Teachers with children, no movers]{Resulting clusters of the teacher dataset with movers eliminated.}\label{TABLE:teacher_children}
%\end{table}

One of the more interesting results that spectral clustering produced on the teacher data was the case where one of two clusters contained only teachers who left.  In this run, movers were removed and only teachers without children were considered.  Spectral clustering grouped $173$ teachers together, all who had quit.  They were all married and had household incomes in the $\$60{,}000$ to $\$74{,}999$ range.  The teachers in this group expressed at most one dissatisfaction in the survey.  In Table \ref{TABLE:teacher_nochildren}, this group is labeled Cluster 1.  Cluster 2 consists of all other teachers that were not movers and did not have children.  Among the $1{,}598$ teachers in Cluster 2, 810 quit.  In a two-proportion $Z$-test, this gave us a one-tailed $p$-value of less that 0.0001, which supports the idea that teachers similar to those in Cluster 1 are more likely to quit than those similar to teachers in Cluster 2.

%\begin{table}[ht]
%\begin{center}
%\begin{tabular}{|l|cc|}
%\hline
%Status & Cluster 1 & Cluster 2 \\ \hline
%Stayer & 0 & 788 \\
%Leaver & 173 & 810 \\ \hline
%Total & 173 & 1,598 \\ \hline
%\end{tabular}
%\end{center}
%\caption[Teacher without children, no movers]{Resulting clusters of the teacher dataset with teachers with children and movers eliminated.}\label{TABLE:teacher_nochildren}
%\end{table}

\begin{table}
\begin{center}
\begin{tabular}{|l|cc|}
\hline
Status & Cluster 1 & Cluster 2 \\ \hline
Stayer & 92 & 1,666\\
Mover & 24 & 1,016\\
Leaver & 81 & 1,649\\ \hline
Total & 197 & 4,331\\ \hline
\end{tabular}
%\end{center}
%\caption[Teacher dataset, all]{Resulting clusters of the entire teacher dataset.}\label{TABLE:teacher_all}
\quad
\begin{tabular}{|l|cc|}
\hline
Status & Cluster 1 & Cluster 2 \\ \hline
Stayer & 100 & 870 \\
Leaver & 48 & 699 \\ \hline
Total & 148 & 1,569 \\ \hline
\end{tabular}
%\end{center}
%\label{TABLE:teacher_children}
\quad
%\subfloat[Resulting clusters of the teacher dataset .]{
\begin{tabular}{|l|cc|}
\hline
Status & Cluster 1 & Cluster 2 \\ \hline
Stayer & 0 & 788 \\
Leaver & 173 & 810 \\ \hline
Total & 173 & 1,598 \\ \hline
\end{tabular}
\end{center}
\caption{Resulting clusters of the entire teacher dataset (left), with movers eliminated (center) and with teachers with children and movers eliminated (right). }\label{TABLE:teacher_nochildren}
\end{table}

To ensure that spectral clustering worked and would give us accurate clusters, the clustering algorithm was applied to a $1/3$ sample of a subgroup of teachers.  Results were used to try to predict the remaining $2/3$.  For example, in the case of the $488$ unmarried teachers, $163$ teachers were sampled (Table \ref{TABLE:teacher_unmarried}).  Spectral clustering grouped $23$ of the teachers in one group because they all did not express complaints and did not have children or other dependents.  Among this first cluster, $17$ had quit while $38$ of the $140$ teachers in the second cluster quit.  This yielded a one-tailed $p$-value of less than 0.0001 where teachers in the first cluster are more likely to quit.  Going through the $325$ unsampled teachers, $55$ displayed the same characteristics as the first cluster.  Proportionally, our prediction that $41$ of the $55$ teachers would quit was not bad considering that in actuality $38$ teachers quit.  Our prediction for the second group of teachers was further off, but still supported the finding that teachers similar to those in the first cluster are more likely to quit than teachers similar to those in the second cluster.  Spectral clustering was able to group the $55$ teachers together in its run with the remaining $2/3$ data points.

\begin{table}[ht]
\begin{center}
\begin{tabular}{|r|c|c|}
\hline
 & With Children Cluster & Without Children Cluster \\ \hline
SC on 1/3 & 17/23 & 38/140 \\ \hline
Predicted & 41/55 & 73/270 \\ \hline
SC on 2/3 & 38/55 & 94/270 \\ \hline
\end{tabular}
\end{center}
\caption[Prediction for unmarried teachers, no movers]{Prediction given by spectral clustering for the teacher dataset (married teachers and movers eliminated).}\label{TABLE:teacher_unmarried}
\end{table}

Although \cite{perona2004self} recommended using the $7^\text{th}$ nearest neighbor to help determine the similarity bandwidth of each point, using different values of nearest neighbor for this dataset yielded different clusters that provided valuable information.  We applied spectral clustering to married teachers who were not movers using the $7^\text{th}$, $50^\text{th}$, and $100^\text{th}$ nearest neighbor.  All teachers in the smaller cluster (Cluster A) using the $7^\text{th}$ nearest neighbor were found in the same cluster (with other teachers) when using the $100^\text{th}$ nearest neighbor.  That same cluster, with the newly added teachers, in the $100^\text{th}$ nearest neighbor was also found in the same cluster with almost the rest of the teachers when using the $50^\text{th}$ nearest neighbor.  Define Cluster B as teachers grouped with teachers in Cluster A using the $100^\text{th}$ nearest neighbor.  Define Cluster C as teachers grouped with teachers in Cluster A and B using the $50^\text{th}$ nearest neighbor.  The remaining teachers make up Cluster D.  A summary of the characteristics of teachers in each cluster can be found on Table \ref{TABLE:Dcluster_summary}.  Note that the table does not reflect correlations; for example, a dissatisfaction value of 1 only appears in Cluster B if the teacher does not have a child.

We found that the difference between all four clusters is statistically significant with a p-value of less than 2.2E-16.  With this, we can obtain a ranking of the teachers where teachers with 2 or 3 points of dissatisfaction are most likely to resign.  Teachers with a young only child and teachers without children, excluding those with the exact characteristics of Cluster A, are very likely to resign.  Teachers with the exact characteristics of Cluster A could possibly resign, while all other teachers with at most one dissatisfaction point are not likely to resign.  In summary, a ranking of teachers who are most likely to quit teaching is achievable with spectral clustering.  It brings us to consider multiple clusters in spectral clustering.

\begin{table}[ht]
\begin{center}
\begin{tabular}{|l|cccc|}
\hline
Characteristics & Cluster A & Cluster B & Cluster C & Cluster D\\ \hline
Household Income & $60{,}000$--$74{,}999$ & Varies & Varies & Varies\\
Children & None & 0 to 1 & Varies & Varies \\
Youngest Child & -1 & -1 to 4 & Varies & Varies \\
Other Dependents & None & None & 0 to 1 & 0 to 3 \\
Dissatisfaction & None & 0 to 1 & 0 to 1 & 2 to 3 \\ \hline
Stayer & 92 & 398 & 772 & 0 \\
Leaver & 81 & 567 & 480 & 214 \\ \hline
Total & 173 & 965 & 1,252 & 214\\ \hline
\end{tabular}
\end{center}
\caption[Married teachers in 4 clusters, no movers]{Four clusters given by altering the tuning parameter on the teacher dataset (unmarried teachers and movers eliminated).} \label{TABLE:Dcluster_summary}
\end{table}

Our analyses of spectral clustering on the teacher data was facilitated by looking at subgroups of teachers.  It is perhaps that the variables that create the divide for the subgroups (i.e.\ unmarried teachers only, teachers without children only, etc.) interact with other variables, such as age of youngest child.  Variable transformations, interaction terms, multiple clusters, and weighting are thus important points of consideration when working with spectral clustering on attrition-like data.

\subsection{Approximation Results}
To measure the effectiveness of the approximation algorithms on the teacher dataset, we ran each one given a different sample size $10$ times and compared average run time and error rate. The ``movers'' category was removed for simplicity.  As the ground truth, we use the answers obtained by the exact spectral clustering algorithm. In other words, we measure the ability of the approximation algorithms to give the same answer as exact spectral clustering in a shorter amount of time (the exact algorithm ran in $476.47$ seconds).

\begin{figure}[ht]
\begin{center}
\includegraphics[width=0.5\textwidth]{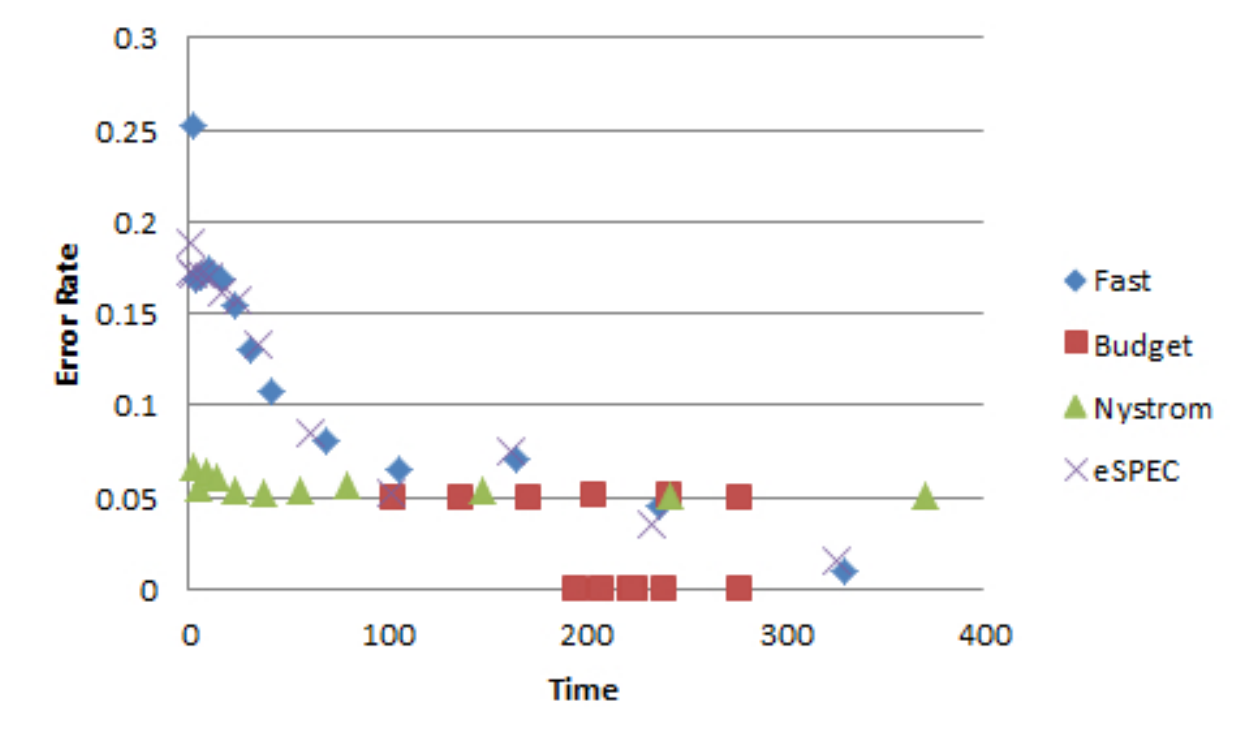}
\caption[Teacher Data (without Movers) Error Rate vs.\ Time Graph]{Error rate versus time (in seconds) plot for the teacher data (without movers).}\label{FIGURE:teachertradeoff}
\end{center}
\end{figure}

\begin{figure}[ht]
\centering
\subfigure[]{
\includegraphics[width=0.45\textwidth]{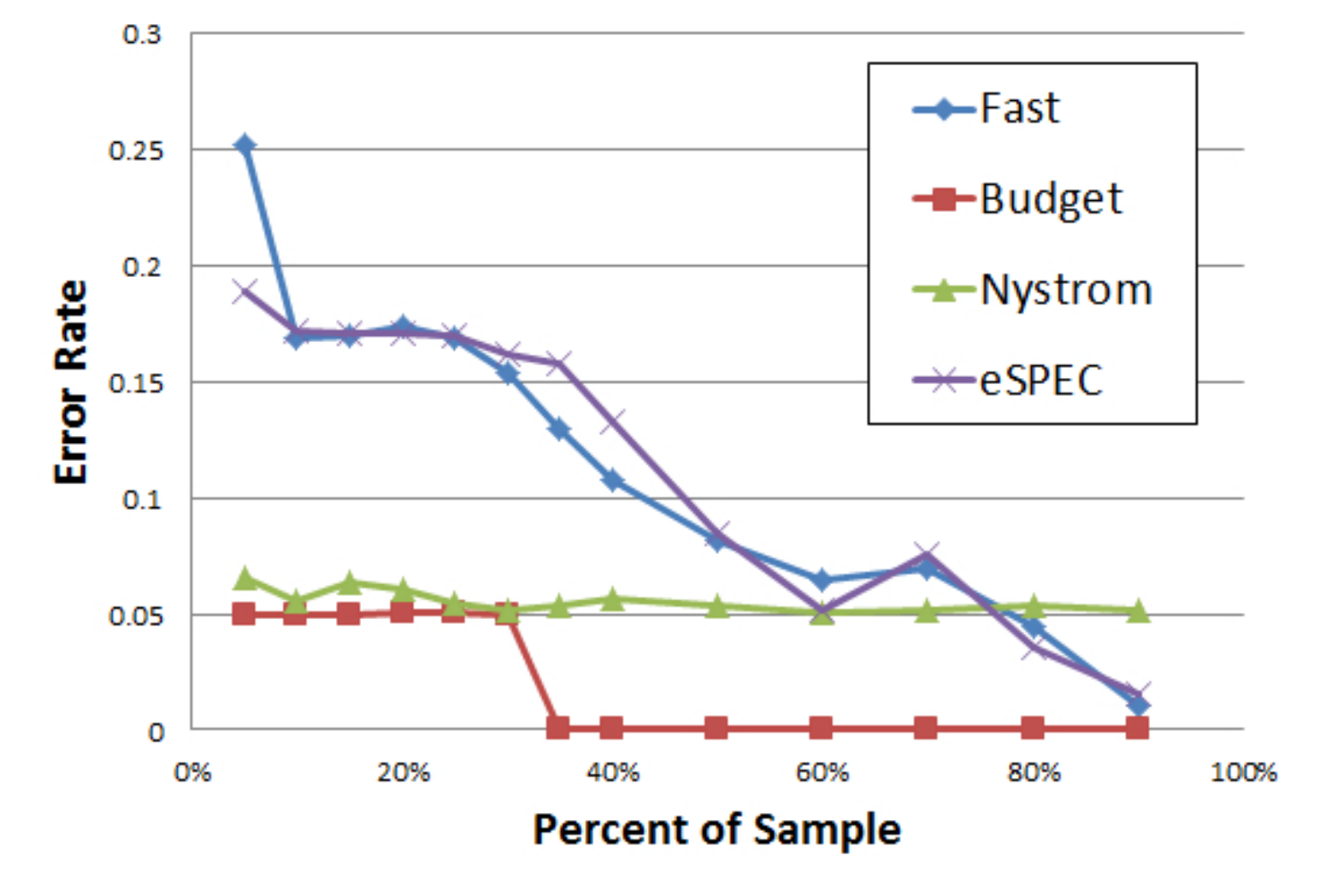}
}
\subfigure[]{
\includegraphics[width=0.45\textwidth]{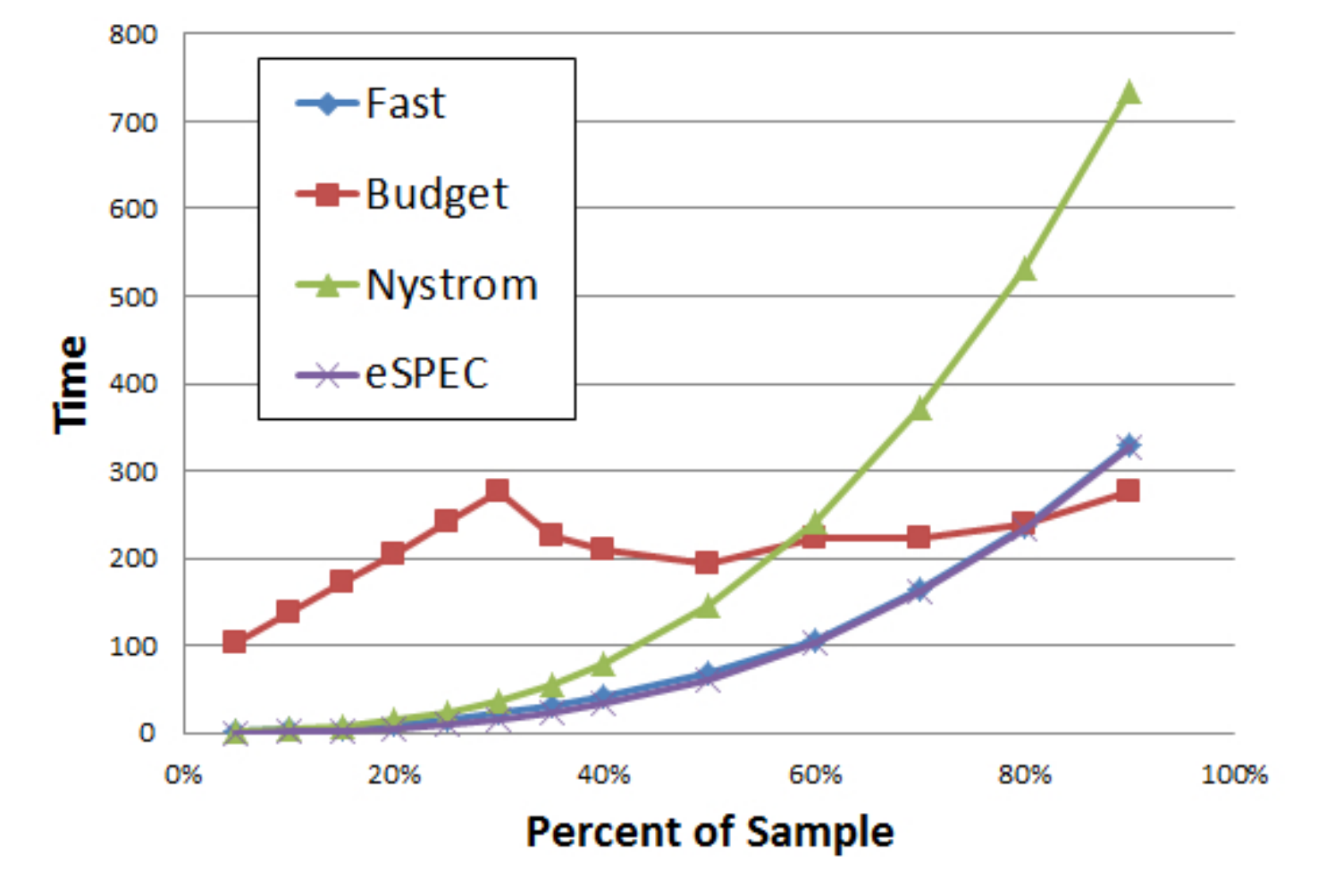}
}
\caption[Teacher Accuracy and Efficiency]{Sample sizes range from 5--100\% of the teacher dataset with $n=3,488$, and without the classification of movers.}\label{FIGURE:teachersresults}
\end{figure}

Although the results are similar to those obtained using visually apparent clusters, we see two major differences. First, as seen in Table~\ref{tab:teachers}, spectral clustering on a budget was not necessarily the slowest or most inaccurate algorithm of the group.  Secondly, as seen in Figure \ref{FIGURE:teachersresults}, it took a longer time for the algorithms to reach zero error.  This is potentially due to the use of proximity of the clusters. Perhaps spectral clustering on a budget handles less structured clusters better than the others.  Still, it displays an odd progression in terms of run time -- one that is not entirely upward sloping, as seen in Figure \ref{FIGURE:teachersresults}.  This leads us to believe it may be unstable in this setting. For this kind of dataset, fast spectral clustering or eSPEC may offer more advantages.  Alternatively, if it is acceptable for the error rate to be up to $5\%$, Nystr\"{o}m gives adequate results the quickest.  A depiction of the tradeoff between efficiency and accuracy for this dataset is given in Figure \ref{FIGURE:teachertradeoff}.

\begin{table}[ht]
\small
\begin{center}
\begin{tabular}{|l|l|l|l|l|l|l|l|l|}
\hline
($n=3{,}488$) & \multicolumn{2}{c|}{Fast}  & \multicolumn{2}{c|}{Budget}  & \multicolumn{2}{c|}{Nystr\"{o}m} & \multicolumn{2}{c|}{eSPEC}\\ \hline
Sample Size & Time & Error & Time & Error & Time & Error & Time & Error\\ \hline
5\% & 1.9906 & 0.2518 & 102.9295 & 0.0499 & 1.7644 & 0.066 & 0.4243 & 0.1889\\ \hline
10\% & 3.4086 & 0.1691 & 136.6257 & 0.0499 & 4.1699 & 0.0551 & 0.8143 & 0.1721\\ \hline
%15\% & 5.8142 & 0.17 & 171.2579 & 0.0499 & 8.1807 & 0.0637 & 1.9235 & 0.1714\\ \hline
20\% & 10.3694 & 0.1735 & 204.2521 & 0.0502 & 14.2835 & 0.0601 & 5.4663 & 0.1709\\ \hline
%25\% & 15.9371 & 0.169 & 242.2072 & 0.0505 & 23.5967 & 0.0543 & 9.9341 & 0.1704\\ \hline
30\% & 22.8323 & 0.1541 & 276.855 & 0.0493 & 37.1969 & 0.0517 & 16.3941 & 0.1618\\ \hline
%35\% & 31.2782 & 0.1299 & 225.6554 & 0 & 56.3164 & 0.0532 & 24.2426 & 0.1577\\ \hline
40\% & 41.5103 & 0.1079 & 208.5421 & 0 & 80.0753 & 0.0567 & 34.2765 & 0.1328\\ \hline
50\% & 69.0788 & 0.0812 & 194.2056 & 0 & 146.9904 & 0.0536 & 61.5502 & 0.0843\\ \hline
60\% & 106.2819 & 0.065 & 222.941 & 0 & 241.4677 & 0.0506 & 102.1619 & 0.052\\ \hline
70\% & 164.4188 & 0.07 & 223.8302 & 0 & 371.103 & 0.0511 & 162.096 & 0.0752\\ \hline
80\% & 236.8828 & 0.045 & 239.8047 & 0 & 532.3269 & 0.0533 & 232.819 & 0.035\\ \hline
90\% & 329.4538 & 0.01 & 277.4946 & 0 & 733.4184	 & 0.0511 & 326.2121 & 0.015\\
\hline
\end{tabular}
\caption[Teachers simulation details]{The run time and error rate of each sample size for each approximation algorithm, ran on the teacher dataset.}
\label{tab:teachers}
\end{center}
\end{table}

\section{Discussion}\label{sec:discuss}
Fast spectral clustering frequently gives the most accurate results in the shortest running time for small datasets using a small $k$. For easily clustered data, this may be due to the $k$-means algorithm overpowering the fast spectral clustering algorithm for really small $k$ values. The Nystr\"{o}m method often performs quickly and accurately as well, especially on the larger or more complicated datasets. eSPEC is the fastest when $n$ is extremely large, but also the most inaccurate.

Intuitively, we sacrifice accuracy for efficiency when we run the approximation algorithms on a relatively small set of points compared to the dataset size.  However, the trend is not so apparent for spectral clustering on a budget.  The other algorithms face approximately the expected tradeoff. Across all datasets, spectral clustering on a budget often takes longer than the other three with limited advance in accuracy.  Interestingly, it consistently reaches a point where smaller sample sizes actually make it increase in running time. Thus, it may be preferable to utilize one of the other three algorithms, depending on the size of the data and the goal of the clustering results.

In addition, when given the right data, spectral clustering can find similarities in individuals that may point to employees at high risk of attrition.  Using a subset of the teacher dataset, we found we could predict with some accuracy which teachers had left.  This shows that if possible, breaking the data down and running spectral clustering on smaller groups is very useful.  Approximation methods did not perform quite as well on the teacher dataset, but they do give accurate results and cut down run time by a few minutes.  If run on a larger employee dataset, they would likely increase efficiency by a greater factor.  With finer tuning of parameters and variable choices, even more improvements may be possible.

\section*{Acknowledgements}
We would like to thank our advisors Christos Boutsidis and Deanna Needell.
We also thank Mike Raugh, Stacey Beggs, Dimi Mavalski, and all the faculty
at the Institute of Pure and Applied Mathematics for directing and coordinating
this summer. Lastly, thank you to IBM and NSF for funding this project.

\bibliographystyle{plain}
\bibliography{Biblio}
\end{document}